\documentclass[preprint,9pt,authoryear]{elsarticle}

\usepackage[T5]{fontenc}
\usepackage{amsmath}
\usepackage{subfigure}
\usepackage{multirow}
\usepackage{url}
\usepackage{fancyhdr}
\usepackage{color}
\usepackage{latexsym}
\usepackage{tablefootnote}
\usepackage{colortbl}
\usepackage{graphicx}
\usepackage{caption}
\usepackage{makecell}
\usepackage{graphicx}%
\usepackage{multirow}%
\usepackage{amssymb,amsfonts}%
\usepackage{amsthm}%
\usepackage{mathrsfs}%
\usepackage[title]{appendix}%
\usepackage{xcolor}%
\usepackage{textcomp}%
\usepackage{manyfoot}%
\usepackage{booktabs}%
\usepackage{algorithm}%
\usepackage{algorithmicx}%
\usepackage{algpseudocode}%
\usepackage{listings}%
\usepackage{microtype}%
\usepackage{tabularray}
\usepackage{array}
\usepackage{lipsum}
\usepackage{float}
\usepackage{adjustbox}
\usepackage[T5]{fontenc} 
\usepackage{subfigure}
\usepackage{amsmath}
\usepackage{adjustbox}
\usepackage{pifont}
\usepackage{tabularx}
\usepackage{pgfplots}
\usepackage{pgfplotstable}
\usepackage{arydshln}
\usepackage{tikz}
\usepackage{collcell}
\usepackage{threeparttable}
\usepackage{graphicx}
\usepackage{longtable}
\usepackage{algorithm}
\usepackage{algpseudocode}
\usepackage{times}
\usepackage{latexsym}
\usepackage{graphicx}
\usepackage{multirow}
\usepackage{pgf-pie}   
\usepackage{soul}
\usepackage{color}
\usepackage{tablefootnote}
\usepackage{adjustbox}
\usepackage{arydshln}
\usepackage{url}
\usepackage{float}
\usepackage{graphicx,graphics}
\usepackage{caption}
\usepackage{subcaption}
\usepackage{amsmath}
\usepackage{subfigure}
\usepackage{multirow}
\usepackage{url}
\usepackage{fancyhdr}
\usepackage{color}
\usepackage{latexsym}
\usepackage{tablefootnote}
\usepackage{colortbl}
\usepackage{xcolor}
\usepackage{graphicx}
\usepackage{caption}
\usepackage{makecell}
\usepackage{graphicx}%
\usepackage{multirow}%
\usepackage{amssymb,amsfonts}%
\usepackage{amsthm}%
\usepackage{mathrsfs}%
\usepackage[title]{appendix}%
\usepackage{xcolor}%
\usepackage{textcomp}%
\usepackage{manyfoot}%
\usepackage{booktabs}%
\usepackage{algorithm}%
\usepackage{algorithmicx}%
\usepackage{algpseudocode}%
\usepackage{listings}%
\usepackage{microtype}%
\usepackage{tabularray}
\usepackage{array}
\usepackage{lipsum}
\usepackage{float}
\usepackage{adjustbox}
\usepackage[T5]{fontenc} 
\usepackage{subfigure}
\usepackage{amsmath}
\usepackage{adjustbox}
\usepackage{pifont}
\usepackage{tabularx}
\usepackage{pgfplots}
\usepackage{pgfplotstable}
\usepackage{arydshln}
\usepackage{tikz}
\usepackage{collcell}
\usepackage{threeparttable}
\usepackage{graphicx}
\usepackage{longtable}
\usepackage{silence}
\usepackage{lipsum}

\usepackage{xcolor}
\usepackage{soul, soulutf8}

\begin{document}
\begin{frontmatter}

\let\WriteBookmarks\relax
\def\floatpagepagefraction{1}
\def\textpagefraction{.001}

\title{Evaluating Large Language Model Capability in Vietnamese Fact-Checking Data Generation}                      

\author[1,3]{Long Truong To}
\ead{21521101@gm.uit.edu.vn}

\author[1,3]{Hung Tuan Le}
\ead{21520250@gm.uit.edu.vn}

\author[2,3]{Dat Van-Thanh Nguyen}
\ead{20520436@gm.uit.edu.vn}

\author[1,3]{Manh Trong Nguyen}
\ead{21520343@gm.uit.edu.vn}

\author[1,3]{Tri Thien Nguyen}
\ead{21522707@gm.uit.edu.vn}
\author[1,3]{Tin Van Huynh\corref{cor1}}
\ead{tinvh@uit.edu.vn}

\author[1,3]{Kiet Van Nguyen\corref{cor1}} 
\ead{kietnv@uit.edu.vn}

\affiliation[1]{organization={Faculty of Information Science and Engineering, University of Information Technology\\},
    city={Ho Chi Minh City},
    country={Vietnam}}

\affiliation[2]{organization={Faculty of Computer Science, University of Information Technology},
    city={Ho Chi Minh City},
    country={Vietnam}}
    
\affiliation[3]{organization={Vietnam National University},
    city={Ho Chi Minh City},
    country={Vietnam}}

\cortext[cor1]{Corresponding author at the University of Information Technology, Vietnam National University, Ho Chi Minh City, Vietnam.}

\begin{abstract}
Large Language Models (LLMs), with gradually improving reading comprehension and reasoning capabilities, are being applied to a range of complex language tasks, including the automatic generation of language data for various purposes. However, research on applying LLMs for automatic data generation in low-resource languages like Vietnamese is still underdeveloped and lacks comprehensive evaluation. In this paper, we explore the use of LLMs for automatic data generation for the Vietnamese fact-checking task, which faces significant data limitations. Specifically, we focus on fact-checking data where claims are synthesized from multiple evidence sentences to assess the information synthesis capabilities of LLMs. We develop an automatic data construction process using simple prompt techniques on LLMs and explore several methods to improve the quality of the generated data. To evaluate the quality of the data generated by LLMs, we conduct both manual quality assessments and performance evaluations using language models. Experimental results and manual evaluations illustrate that while the quality of the generated data has significantly improved through fine-tuning techniques, LLMs still cannot match the data quality produced by humans.
\end{abstract}
\end{frontmatter}


\section{INTRODUCTION}
The rise of misinformation online has spurred the development of fact-checking tasks, which verify misleading information using multiple knowledge sources \cite{politifact}. Creating an automatic fact-checking system demands extensive training data \cite{thorne-etal-2018-fever,VitaminC}, posing a challenge for low-resource languages like Vietnamese due to limited datasets. Efforts to build Vietnamese fact-checking datasets include ViWikiFC \cite{viwikifc}, and ISE-DSC01\footnote{https://dsc.uit.edu.vn/}, derived from news sources. Despite these efforts, there remains a need for high-quality, diverse data to advance this task.

Large language models (LLMs) trained on vast datasets are crucial in natural language processing, excelling in context understanding and real-world tasks \cite{sawada2023arb,valmeekam2022large,guha2024legalbench}. Furthermore, LLMs have been used for automatic data construction recently, reducing costs compared to manual methods \cite{goel2023llms,dai-etal-2023-gpt}. However, this application focuses on English predominantly, with languages like Vietnamese remaining underexplored.

\begin{figure*}[t]
    \centering
    \includegraphics[width=1\textwidth]{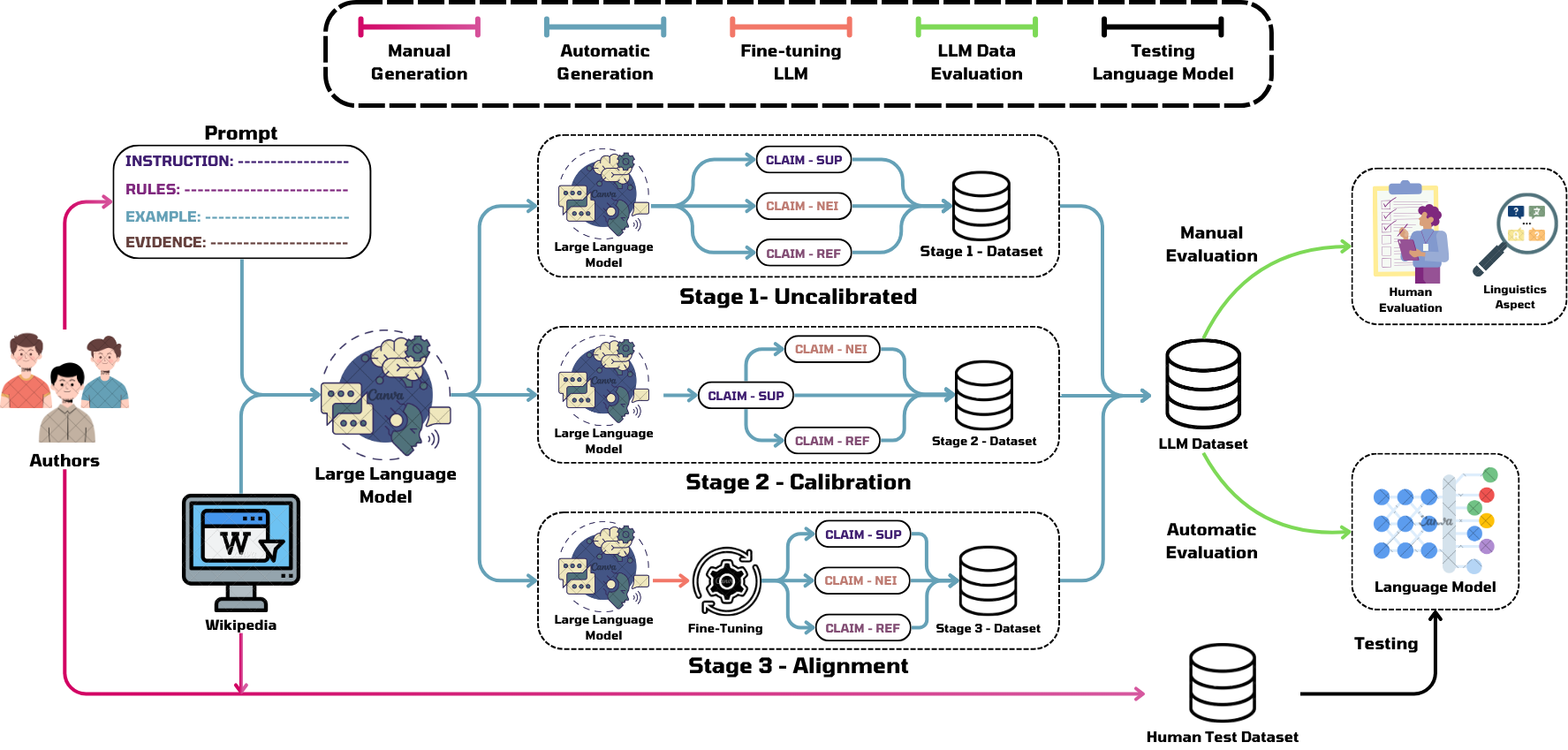}
    \caption{Large language model data generation process. Begin standard prompt construction and evidence selection from Wikipedia (see Section \ref{subsec:Evidence Selection}). In the automatic data generation phase through LLM, we conduct three stages of generation, including using the standard prompt in Stage 1 - Uncalibrated (see Section \ref{subsec:Prompting Experimental Result}), Stage 2 - Calibration with modified prompt and different flow (see Section \ref{subsubsec:Why do we need to fine-tune LLMs in the generated task?}), finally stage 3 - alignment we are fine-tuning LLM and using the standard prompt for generation (see Section \ref{subsubsec:Results Of Fine-Tuning Stage}). After data generation, we evaluate LLM dataset quality through language model (see Section \ref{sec:AUTOMATIC EVALUATION}) and human evaluation (see Section \ref{subsec:MANUAL EVALUATION}).} 
    \label{fig:Large language model data generation process}
\end{figure*}

To address the challenges in Vietnamese fact-checking, we undertake this study to meet the demand for Vietnamese fact-checking data and to explore the role of large language models (LLMs) in constructing non-English data. We design an automatic data construction process using LLMs (see Figure \ref{fig:Large language model data generation process}). Consistency is ensured by developing a standard prompt (see Section \ref{subsubsec:Prompt Preparation}) using techniques like instruction prompting and few-shot learning. We generate data with this prompt on five LLMs: four multilingual Llama2 \cite{llama2}, Qwen \cite{qwen}, Gemini \cite{gemini}, and GPT-3.5\footnote{https://openai.com/}, and one monolingual Vistral\footnote{https://huggingface.co/Viet-Mistral/Vistral-7B-Chat}. Unlike other Vietnamese fact-checking datasets, our LLM-generated data constructs each claim from multiple evidence sentences, allowing us to evaluate the ability of LLMs to synthesize and combine information.

We evaluate the quality of LLM-generated data using pre-trained language models on a custom test set. These models include multilingual ones like XLM-R \cite{xlmr}, mBERT \cite{bert}, and mDeBERTaV3 \cite{he2022debertav3}, as well as the Vietnamese monolingual model PhoBERT \cite{nguyen-tuan-nguyen-2020-phobert}. In addition to using language models, we perform a manual evaluation based on established criteria. Finally, we analyze linguistic features such as new word rate and overlap in the LLM-generated data.

Our article contributions include:
\begin{itemize}
    \item We have developed new automatically generated Vietnamese fact-checking datasets that abstract information from multiple pieces of evidence to create each claim. Unlike existing Vietnamese fact-checking datasets, which usually rely on claims derived from a single piece of evidence, our datasets are uniquely constructed by synthesizing multiple evidence sentences for each claim. By using large language models (LLMs) in the data generation process, we significantly reduce the time and cost associated with traditional manual annotation.
    \item We are pioneers in evaluating large language models (LLMs) for generating Vietnamese fact-checking data, conducting extensive manual and automatic evaluations. Our study includes an assessment of five LLMs: four multilingual models (Llama2, Qwen, Gemini, GPT-3.5) and a Vietnamese monolingual model (Vistral). We employ automatic evaluations using advanced language models like XLM-R, mBERT, mDeBERTaV3, and PhoBERT on test sets, alongside detailed manual evaluations based on established criteria to assess quality, coherence, and factual accuracy.
    \item We analyze various linguistic aspects of the LLM-generated datasets to gain a deeper understanding of the quality and behavior of LLMs in Vietnamese data generation. Our analysis includes examining linguistic features such as new word rate and textual overlap across models, identifying common error patterns and abnormal cases to pinpoint areas for improvement. We also explore the effectiveness of supervised fine-tuning techniques to enhance LLM performance. 
\end{itemize}

In this study, we organize the paper as follows. In Section \ref{sec:Related Work}, we review research on fact-checking in Vietnamese, as well as studies that explore the application of Large Language Models (LLMs) in automatic data generation. In Section \ref{sec:Data Generation}, we describe the data generation process, including setup and preparation of raw data. We evaluate the quality of generated data using two methods: manual evaluation (see Section \ref{subsec:MANUAL EVALUATION}) and performance-based evaluation through language models (see Section \ref{sec:AUTOMATIC EVALUATION}). Additionally, we apply techniques to enhance the data generation capabilities of LLMs (see Section \ref{sec:Supervised-finetuning LLM}). Finally, in Section \ref{sec:CONCLUSION AND FUTURE WORK}, we summarize the results achieved using LLMs in data generation and suggest directions for future work.

\section{RELATED WORK}
\label{sec:Related Work}
In this section, we review existing work and challenges in Vietnamese NLP, especially focusing on the limitations of current resources and the recent advances in leveraging the prompting technique of Large Language Models (LLMs) and supervised fine-tuning techniques to address these gaps such as Low-Rank Adaptation (LoRA) for efficient model adaptation.

\subsection{Insufficient Resources}
\label{subsec:Insufficient Resources}
    The development of NLP research on Vietnamese faces significant challenges due to limited resources \cite{ngo2019overcoming,ngo2020improving,vigptqa}. Acknowledging these limitations, efforts have been made to develop handcrafted datasets for various tasks, such as ViWikiFC \cite{viwikifc} and ISE-DSC\footnote{https://dsc.uit.edu.vn/} for fact-checking, and ViNLI \cite{huynh-etal-2022-vinli} and ViHealthNLI \cite{nguyen-etal-2024-vihealthnli} for natural language inference (NLI). However, the amount of data available remains modest compared to resources in other languages \cite{thorne-etal-2018-fever,hu-etal-2022-chef,VitaminC}. Additionally, the existing datasets for tasks like fact-checking and NLI focus on constructing new sentences with different semantic variations based on a sample sentence, which makes these datasets less challenging for language models.

In this study, we address the aforementioned issues by utilizing Large Language Models (LLMs) to generate additional data for fact-checking tasks. Moreover, our data focuses on synthetic data inference, meaning that each data sample will be generated through the synthesis of information from multiple sentences.

\subsection{The Capability of Large Language Models}
\label{subsec:Large Language Model Capability}
Large language Models (LLMs) possess immense capabilities that are still being fully explored. They are increasingly used in research for essay writing support \cite{yuan2022wordcraft}, coding assistance \cite{nam2024using}, data generation \cite{tang2023does}, and model evaluation \cite{zhou2023don}, including reinforcement learning from human feedback (RLHF) \cite{ouyang2022training,rafailov2024direct}. LLMs like GPT-3, BERT, and their successors have revolutionized natural language processing (NLP) tasks. 

For instance, GPT-3, introduced by \citeauthor{gpt3}, shows remarkable performance in generating human-like text, performing complex language tasks with minimal supervision \cite{kojima2022large}, and adapting to a wide range of applications through few-shot learning \cite{gpt3}. Similarly, BERT, developed by \citeauthor{bert}, is instrumental in achieving state-of-the-art results in various NLP benchmarks \cite{wang2018glue,yang2018hotpotqa} by leveraging bidirectional representations from transformers.

This article focuses on synthetic data inference, an area where LLMs can significantly contribute. Generating high-quality labeled data usually requires substantial effort and resources. We aim to leverage LLMs to create such data efficiently. The process involves inputting a paragraph with multiple sentences and a label, LLMs generate a corresponding claim sentence. This assesses the understanding and inferencing capabilities of models, which is crucial for data augmentation and artificial data synthesis.  
\subsection{Supervised Fine-Tuning}
\label{subsec:Supervised Fine-Tuning}
Supervised Fine-Tuning (SFT) is a crucial technique in machine learning \cite{erhan2010does} that leverages labeled data to adapt pre-trained models for specific tasks \cite{howard2018universal,gunel2020supervised}, improving their accuracy and performance. This section reviews the foundational work, advances, and challenges in SFT, particularly focusing on the concept of Self Fine-Tuning.

The idea of self-improvement in machine learning models, where the model utilizes its own predictions for further training, has been explored in various contexts. Early works by \citeauthor{sft_early} introduce the concept of Noisy Student training, where a student model is trained using both labeled data and pseudo-labels generated by a larger teacher model. This approach demonstrates significant improvements in image classification tasks.

In the domain of natural language processing (NLP), self fine-tuning is applied to improve language models' performance on specific tasks. \citeauthor{sft_nlp} explored ELECTRA, a model that uses replaced token detection as a pre-training task. This method enhances the fine-tuning process by making the model more sensitive to the specifics of the target task, thereby improving its performance on downstream NLP tasks.
\subsection{LoRA Training Model}
\label{subsec:LoRA training model}
Low-Rank Adaptation (LoRA) \cite{lora} is a technique with the ability to optimize in fine-tuning of large language models (LLMs) \cite{liu2022few,li2024pre}. By introducing low-rank matrices to the layers of pre-trained models, LoRA allows for efficient adaptation without updating all the parameters of models \cite{ding2023parameter}.

Initially explored to reduce computational costs, LoRA is formalized by \citeauthor{lora}, who demonstrated that adding trainable low-rank matrices to each transformer layer significantly cuts down the parameters needed for fine-tuning. This method maintains the pre-trained knowledge while enabling efficient task-specific adaptation.

Since its introduction, LoRA has been refined, reducing trainable parameters by 10 to 100 times, as shown by \citeauthor{lora}. This efficiency makes it feasible to fine-tune large models with limited computational resources. LoRA is successfully applied to tasks such as natural language understanding \cite{wu2024infoprompt,ye2023ureader}, machine translation \cite{zhang2023machine,iyer2023towards}, and text generation \cite{li2024pre} by adjusting low-rank matrices to capture task-specific features while preserving the general knowledge of model.

Our research proposes creating a dataset to abstract multi-sentence evidence into a claim sentence, aiming to save time and resources. We utilize LoRA methods for efficiency instead of training the full model.
\section{DATA GENERATION}
\label{sec:Data Generation}

This section outlines the preparation of raw evidence and standard prompts for automatic data generation. It also provides statistics and analysis of the datasets post-generation, focusing on linguistic features like new word rate and overlap between claim and evidence.

\subsection{Evidence Selection}
\label{subsec:Evidence Selection}
We utilize Wikipedia\footnote{https://vi.wikipedia.org/wiki/} as the primary data source, covering various fields and collecting 149 topics, spanning from titles on history, science, and celebrities. From these topics, we randomly extract two to three sentences from the same paragraph within each topic to ensure the coherence of the gathered evidence. This is different from the previous experiment, in which this experiment we expect the generated claim to be abstract from the evidence paragraph. As a result, we obtained 7,286 evidence sentences corresponding to 3,643 data lines.

\subsection{Automatic Data Generation}
\label{subsec:Automatic Data Generation}
\subsubsection{Prompt Preparation}
\label{subsubsec:Prompt Preparation}
We apply few-shot learning with clear instructions for each label to ensure accurate task comprehension \cite{liu2023pre,reynolds2021prompt}. First, we define the role of model \cite{liu2023pre} and explain the \textit{\textit{Supported}}, \textit{Refuted}, and \textit{Not\_Enough\_Information} (\textit{NEI}) labels \cite{huang2023not}. Guidelines ensure pieces of evidence are abstracted without full extraction \cite{reynolds2021prompt}, and generated information for \textit{\textit{Supported}} and \textit{Refuted} labels stays within the evidence. Responses include only a claim, guided by examples and the "[CLAIM]" token \cite{bsharat2023principled}.

In the calibration stage of data generation, we maintain the same rule setup but introduce specific variations. For sentences with \textit{Refuted} label, we provide models with the \textit{\textit{Supported}} claim for the same pieces of evidence. For \textit{NEI} labels, the model receives both \textit{\textit{Supported}} and \textit{Refuted} claims related to the evidence. This approach ensures nuanced and accurate sentence generation across different labels. For each dataset generated, we sample a random hundred samples to check by hand the fails rule and apply it to all samples in the same dataset (Section \ref{appendix:Abnormal cases in generating}). 

\subsubsection{Generation-Config Setup}
\label{subsec:Generation-config setup}
The settings we use for models are the same, but we set different parameters for distinct labels. 
We use "top\_p" for all labels with 0.7 and "top\_k" is 10. On the other hand, for \textit{\textit{Supported}} and \textit{Refuted} labels, we assign lower "temperature" with 0.5 and 0.4 respectively to encourage models to generate within the bounds of evidence while penalizing the emergence of unexpected tokens. In the \textit{NEI} label, we need to stimulate the creativity and integration of models for external information so that we opt for a higher "temperature" of 0.9.
\subsubsection{Generating Data With LLMs}
\label{subsec:Generating data with LLMs}
After setting up all the prompts and parameters, we provide the LLMs with the extracted evidence, as outlined in Section \ref{subsec:Evidence Selection}. The goal is to produce a claim that meets all the requirements specified in the prompt and aligns with the desired label. The entire process is illustrated in Figure \ref{fig:Large language model data generation process}, where five models are used during the uncalibrated phase, followed by three models in both the calibration and alignment phases. 

The resulting dataset from each model is then divided into three subsets: train, dev, and test. In addition to the test set derived from the LLM-generated data, we also construct a separate test set manually curated by humans. This human-generated test set serves as a benchmark to evaluate the performance of the models trained on the datasets produced by the LLMs.

\subsubsection{Abnormal Cases in Generating}
\label{appendix:Abnormal cases in generating}

In the task we define the model to generate the claims from the pieces of evidence, we notice that:
\begin{itemize}
    \item The model generates by the way answers our questions, not carries out orders so that the model generates the "CLAIM" and "hope" again in the output. In these cases, it seems that LLMs cannot understand our prompt and this happens most when using Llama2 (Table \ref{tab: Comprehension}). \\
    Ex: "Hope you create the CLAIM based on the provided EVIDENCE!" or "Hope this works!".
\begin{table}[htb]
\scriptsize
\centering
\begin{tabular}{p{0.2cm}|p{7cm}|p{2.5cm}}
\hline
\multicolumn{1}{p{1.4cm}|}{\textbf{Category}} & \multicolumn{1}{c|}{\textbf{Evidence}} & \multicolumn{1}{c}{\textbf{Claim}} \\ \hline
\multicolumn{1}{p{0.25cm}|}{\textbf{Wrong Cases}} & Đến năm 1902, Hà Nội trở thành thủ đô của toàn Liên bang Đông Dương. Vào năm 1921, toàn thành phố có khoảng 4.000 dân châu Âu và 100.000 dân bản địa. Sự xuất hiện của tầng lớp tư sản Việt Nam khiến văn hóa Hà Nội cũng thay đổi.

(\textit{By 1902, Hanoi had become the capital of the entire Indochinese Federation. In 1921, the city had around 4,000 European residents and 100,000 local residents. The emergence of the Vietnamese bourgeoisie also brought changes to Hanoi's culture.})
& Hope you can create a CLAIM based on the provided EVIDENCE!
 \\ \hline
\end{tabular}
\caption{An example of a claim statement when the model does not understand the intention of the prompt (The claims is generated by model Llama2).}
\label{tab: Comprehension}
\end{table}
    \item The model generates claims extracted from the evidence, rather than abstracting the information from the evidence (Table \ref{tab:Extracted Claim}). 
\begin{table}[htb]
\scriptsize
\centering
\begin{tabular}{p{0.2cm}|p{7cm}|p{2.5cm}}
\hline
\multicolumn{1}{p{1.4cm}|}{\textbf{Category}} & \multicolumn{1}{c|}{\textbf{Evidence}} & \multicolumn{1}{c}{\textbf{Claim}} \\ \hline
\multicolumn{1}{p{0.25cm}|}{\textbf{Wrong Cases}} & Các nhà sử học nhân văn chủ nghĩa biện luận rằng \textcolor{red}{giới học giả đương thời đã khôi phục những liên hệ trực tiếp với thời cổ điển, do đó bỏ qua thời kỳ trung gian}, mà họ lần đầu tiên gọi là thời Trung Đại, mà bấy giờ trong các tài liệu tiếng Ý là media tempestas. Thuật ngữ la rinascita (sự tái sinh, phục hưng) cũng xuất hiện, tuy nhiên, trong nghĩa rộng mà Giorgio Vasari gán cho trong cuốn Vite de' più eccellenti architetti, pittori, et scultori Italiani ("Cuộc đời của các nhà kiến trúc, họa sĩ và nhà điêu khắc Ý tài ba", 1550).

(\textit{Humanist historians argue that \textcolor{red}{contemporary scholars have restored direct links with the classical period, thereby bypassing the intermediate period}, which they first called the classical period. Middle Ages, which at that time in Italian documents was called media tempestas. The term la rinascita (rebirth, revival) also appears, however, in the broader sense given by Giorgio Vasari in his Vite de' più eccellenti architetti, pittori, et scultori Italiani ("Lives of the Architects , talented Italian painter and sculptor", 1550).})
& Giới học giả đương thời đã khôi phục những liên hệ trực tiếp với thời cổ điển, do đó bỏ qua thời kỳ trung gian. 

(\textit{Contemporary scholars have restored direct links with the classical period, thereby bypassing the intermediate period})\\ \hline
\end{tabular}
\caption{An example of a large language model is generated by extracted from evidence.}
\label{tab:Extracted Claim}
\end{table}
    \item The model generates the explanation together with the claim. So we decide to eliminate the explanation and keep the claim.
    \item Models like Qwen and LLama2 have the problem that usually generates multilingual; a lot of claims are half Vietnamese - half English (for Llama2) and half Vietnamese - half Chinese (for Qwen), so we decided to choose a threshold of English with 30\% and 5\% for Chinese. If the words in claims exceed this edge, we also dispose of them from data. To detect the percentage of non-Vietnamese language in the generation dataset, we use cld3\footnote{https://github.com/google/cld3} which is motivated from mC4 dataset \cite{mt5}.
\end{itemize}

\subsection{Data Generated Statistics}
\label{subsec:Data Generated linguistics features statistic}

The automatic data generation results in 10 datasets across all three stages; each dataset is pre-processed to discard any samples with faults and abnormal cases, as detailed in Section \ref{appendix:Abnormal cases in generating}. The statistics of each dataset are described in Table \ref{tab:LLM generated dataset stastistic}.

\begin{table}[htb]
\centering
\footnotesize

\begin{tabular}{c||l|c:c:c|c}
\hline

\multicolumn{1}{c||}{\multirow{2}{*}{\textbf{Stage}}} & \multicolumn{1}{c|}{\multirow{2}{*}{\textbf{Model}}} & \multicolumn{3}{c|}{\textbf{Number of good data}} & \multirow{2}{*}{\textbf{\begin{tabular}{c}
    Proportion of \\ good data (\%)
\end{tabular}}} \\ \cline{3-5}
\multicolumn{1}{c||}{} & \multicolumn{1}{c|}{} & \multicolumn{1}{c:}{\textbf{\textit{Supported}}} & \multicolumn{1}{c:}{\textbf{\textit{Refuted}}} & \textbf{\textit{NEI}} &  \\ \hline
\multirow{5}{*}{\textbf{Uncalibrated stage}} & Vistral & 3,183 & 3,468 & 3,459 & 92.51 \\
 & Gemini & 3,635 & 3,641 & 3,632 & 99.81 \\
 & Qwen & 3,413 & 3,330 & 3,374 & 92.57 \\
 & GPT & 3,617 & 3,608 & 3,573 & 98.80 \\
 & Llama2 & 1,677 & 1,944 & 1,187 & 43.99 \\ \hline
\multirow{3}{*}{\textbf{Calibration stage}} & Vistral & 3,450 & 3,614 & 3,452 & 96.23 \\
 & Qwen & 3,208 & 3,163 & 3,123 & 86.88 \\
 & Gemini & 3,627 & 3,643 & 3,622 & 99.66 \\ \hline
\multirow{2}{*}{\textbf{Alignment stage}} & Vistral & 3,594 & 3,553 & 3,609 & 98.42 \\
 & Gemini & 3,461 & 3,552 & 3,633 & 97.41\\ 
  & Qwen & 3,510 & 2,650 & 3,033 & 84.12\\ \hline
 \textbf{Human} & \multicolumn{1}{c|}{-} & 539 & 539 & 529 & \multicolumn{1}{c}{-}\\
 \hline

\end{tabular}%
\caption{LLM generated dataset stastistic.}
\label{tab:LLM generated dataset stastistic}
\end{table}

Looking at the percentage of good data in the datasets, the rate is generally high. However, Llama2 at uncalibrated stage and Qwen at calibration stage have the lowest percentages of good data, with Llama2 notably at only 43.99\%. This indicates that Llama2 is the most unstable generation among the five models, followed by Qwen, which has a higher rate of Chinese occurrences when there are changes in the prompt at calibration stage.

Additionally, to provide a basis for comparing the quality of the LLMs generated datasets, we create a dataset with claims conducted from multiple evidence sentences, different from the ViWikiFC and ISE-DSC01 datasets. The statistics for the human-created dataset are also presented in the table below.
\begin{table*}[t]
\centering
\resizebox{\textwidth}{!}{%
\begin{threeparttable}
\begin{tabular}{c||c|c|c|c:c:c:c:c:c|c:c}
\hline
\multicolumn{1}{c||}{\multirow{2}{*}{\textbf{Stage}}} & \multirow{2}{*}{\textbf{Model}} & \multirow{2}{*}{\textbf{Label}} & \multirow{2}{*}{\textbf{New word rate (\%)}} & \multicolumn{6}{c|}{\textbf{New Part-Of-Speech (\%)}} & \multicolumn{2}{c}{\textbf{Overlapping}} \\ \cdashline{5-12}
\multicolumn{1}{c||}{} &  &  &  & \textbf{Noun} & \textbf{Verb} & \textbf{Adjective} & \textbf{Preposition} & \textbf{Adjunct} & \textbf{Other} & \textbf{Jaccard (\%)} & \textbf{LCS} \\ \hline
\multirow{15}{*}{\begin{tabular}{c}
   \textbf{Uncalibrated stage} \\
   \textbf{}
\end{tabular}} & \multirow{3}{*}{Vistral} & \textit{Supported} & 9.54 & 28.61 & 31.03 & 7.56 & 13.52 & 5.19 & 14.09 & 35.18 & 91.43 \\ 
 &  & \textit{Refuted} & 17.31 & 23.82 & 25.63 & 11.26 & 9.93 & 13.04 & 16.32 & 31.41 & 87.77 \\
 &  & \textit{NEI} & 37.79 & 32.43 & 27.24 & 9.67 & 11.71 & 7.46 & 11.53 & 24.07 & 82.24 \\ \cdashline{2-12}
 & \multirow{3}{*}{Gemini} & \textit{Supported} & 16.62 & 29.43 & 32.84 & 7.40 & 12.92 & 4.59 & 12.84 & 28.29 & 72.88 \\
 &  & \textit{Refuted} & 20.26 & 28.89 & 27.07 & 9.57 & 9.19 & 9.15 & 16.13 & 22.96 & 60.21 \\
 &  & \textit{NEI} & 29.41 & 31.28 & 27.84 & 11.25 & 11.10 & 7.03 & 11.51 & 20.04 & 60.35 \\ \cdashline{2-12}
 & \multirow{3}{*}{Llama2} & \textit{Supported} & 10.25 & 20.42 & 23.41 & 4.57 & 9.50 & 4.27 & 37.83 & 37.02 & 97.16 \\
 &  & \textit{Refuted} & 12.08 & 20.38 & 22.00 & 8.46 & 11.64 & 10.51 & 27.01 & 38.48 & 99.42 \\
 &  & \textit{NEI} & 24.86 & 28.28 & 24.96 & 9.81 & 10.67 & 7.22 & 19.06 & 29.5 & 81.53 \\ \cdashline{2-12}
 & \multirow{3}{*}{Qwen} & \textit{Supported} & 28.65 & 27.98 & 28.89 & 7.53 & 12.27 & 5.64 & 17.69 & 42.85 & 141.68 \\
 &  & \textit{Refuted} & 33.96 & 20.22 & 24.26 & 7.62 & 9.45 & 16.06 & 22.39 & 32.8 & 106.58 \\
 &  & \textit{NEI} & 49.26 & 30.89 & 25.21 & 9.71 & 11.32 & 5.42 & 17.46 & 29.27 & 135.53 \\ \cdashline{2-12}
 & \multirow{3}{*}{GPT3.5} & \textit{Supported} & 15.88 & 22.85 & 30.55 & 6.84 & 13.13 & 8.42 & 18.21 & 40.43 & 105.89 \\
 &  & \textit{Refuted} & 20.45 & 17.59 & 25.77 & 8.85 & 8.92 & 16.57 & 22.32 & 42.51 & 118.65 \\
 &  & \textit{NEI} & 36.97 & 27.45 & 27.61 & 10.13 & 11.24 & 8.86 & 14.71 & 28.78 & 96.04 \\ \hline
\multirow{9}{*}{\begin{tabular}{c}
   \textbf{Calibration stage}
\end{tabular}} & \multirow{3}{*}{Vistral} & \textit{Supported} & 20.44 & 30.73 & 28.46 & 7.82 & 13.38 & 5.21 & 14.4 & 35.91 & 107.79 \\
 &  & \textit{Refuted} & 26.39 & 26.56 & 24.97 & 11.01 & 11.19 & 11.74 & 14.53 & 31.71 & 100.06 \\
 &  & \textit{NEI} & 33.36 & 28.74 & 25.17 & 11.81 & 12.63 & 8.02 & 13.63 & 30.94 & 105.42 \\ \cdashline{2-12}
 & \multirow{3}{*}{Gemini} & \textit{Supported} & 14.48 & 27.01 & 32.88 & 7.24 & 12.97 & 4.92 & 15.02 & 33.84 & 88.55 \\
 &  & \textit{Refuted} & 24.24 & 20.91 & 25.09 & 10.45 & 9.13 & 17.57 & 16.85 & 27.04 & 77.04 \\
 &  & \textit{NEI} & 28.08 & 30.46 & 27.96 & 8.42 & 11.41 & 7.52 & 14.23 & 26.12 & 77.22 \\  \cdashline{2-12}
 & \multirow{3}{*}{Qwen} & \textit{Supported} & 39.29 & 27.18 & 25.11 & 8.59 & 10.26 & 4.35 & 24.51 & 31.07 & 111.42 \\
 &  & \textit{Refuted} & 45.59 & 21.22 & 23.05 & 7.76 & 8.77 & 9.47 & 29.73 & 25.89 & 98.04 \\
 &  & \textit{NEI} & 55.65 & 25.91 & 21.54 & 11.62 & 10.29 & 7.01 & 23.63 & 22.61 & 111.66 \\ \hline
\multirow{9}{*}{\begin{tabular}{c}
   \textbf{Alignment stage}\tnote{*}
\end{tabular}} & \multirow{3}{*}{Vistral} & \textit{Supported} & 27.74 & 33.04 & 28.37 & 6.93 & 8.58 & 6.59 & 16.49 & 35.44 & 114.49 \\
 &  & \textit{Refuted} & 27.08 & 35.99 & 24.77 & 8.31 & 6.74 & 8.11 & 16.08 & 35.86 & 116.14 \\
 &  & \textit{NEI} & 52.12 & 42.09 & 20.94 & 9.15 & 7.22 & 5.16 & 15.44 & 19.97 & 96.23 \\ \cdashline{2-12}
 & \multirow{3}{*}{Gemini} & \textit{Supported} & 13.77 & 23.31 & 33.79 & 7.58 & 10.79 & 8.54 & 15.99 & 36.32 & 95.59 \\
 &  & \textit{Refuted} & 11.36 & 20.47 & 21.88 & 11.36 & 6.33 & 15.36 & 24.64 & 39.75 & 109.61 \\
 &  & \textit{NEI} & 42.17 & 33.15 & 24.81 & 10.52 & 9.27 & 8.23 & 14.09 & 21.78 & 78.69 \\ \cdashline{2-12}
  & \multirow{3}{*}{Qwen} & \textit{Supported} & 22.32 & 30.01 & 25.10 & 7.68 & 8.67 & 8.96 & 19.58 & 42.70 & 137.80 \\
 &  & \textit{Refuted} & 22.79 & 24.42 & 25.54 & 8.94 & 7.46 & 14.77 & 18.87 & 34.85 & 108.98 \\
 &  & \textit{NEI} & 46.21 & 34.69 & 23.54 & 9.67 & 8.49 & 8.22 & 15.39 & 23.95 & 98.36 \\ \cdashline{2-12}
 \hline
\multirow{3}{*}{\textbf{Human}} & \multirow{3}{*}{\textbf{-}} & \textit{Supported} & 32.99 & 29.97 & 25.78 & 8.21 & 10.22 & 8.34 & 17.51 & 28.28 & 97.24 \\
 &  & \textit{Refuted} & 33.24 & 29.39 & 24.58 & 8.42 & 9.28 & 9.21 & 19.12 & 26.6 & 88.83 \\
 &  & \textit{NEI} & 39.79 & 32.18 & 25.05 & 8.05 & 9.79 & 7.70 & 17.23 & 25.45 & 94.61 \\
\hline
\end{tabular}
\begin{tablenotes}
\item[*]At stage 3, we only do statistics on the dataset of synthetic training.
\end{tablenotes} 
\end{threeparttable}%
}
\caption{LLM-Generated Dataset Linguistics features Statistics.}
\label{tab:LLM-Generated Dataset Linguistics features Statistics Claim-Evidence}
\end{table*}
 
As mentioned at the beginning of section \ref{sec:Data Generation}, we analyze linguistic features to evaluate the generative capabilities of LLMs through LLM-generated datasets using the two linguistic features below:

\textbf{New Word Rate:} In Vietnamese, words are often composed of multiple syllables, which makes word segmentation an essential step in processing the language. For example, the word "học sinh" (meaning "student") consists of two syllables, "học" and "sinh," that together form a single word with a specific meaning. Another example is "giáo viên" (meaning "teacher"), where "giáo" and "viên" are two syllables that combine to create a single word. Unlike some languages where each syllable is typically part of a distinct word, a lot of words in Vietnamese rely on combinations of syllables to convey complete concepts, so we use VnCoreNLP \cite{vu-etal-2018-vncorenlp} to segment the data from syllables into words which provides more accurate linguistic statistics. The statistics results are presented in Table \ref{tab:LLM-Generated Dataset Linguistics features Statistics Claim-Evidence}.

Overall, the new word rate among the three labels across all 11 datasets generated by LLM is unbalanced, with the highest rate in the \textit{NEI} label and the lowest in the \textit{\textit{Supported}} label. This indicates that LLMs exhibit high creativity when generating the \textit{NEI} label; however, this high rate does not correspond with external information outside pieces of evidence, thus creating claims that adhere to the requirements of \textit{NEI} label, as we explain in Section \ref{subsubsec:Why do we need to fine-tune LLMs in the generated task?}. Additionally, we analyze the generation tendencies of LLMs through the part-of-speech of new words in claims, referencing the human evaluation to explain why the \textit{NEI} label has high linguistic features. Table \ref{tab:LLM-Generated Dataset Linguistics features Statistics Claim-Evidence} shows that LLMs tend to generate new nouns. Furthermore, for the \textit{Refuted} label, the rate of new adjuncts is always higher than the other two labels across the 11 LLM-generated datasets. This is understandable as adjuncts in \textit{Refuted} sentences often include negative words like "không", "chưa", "chẳng" (not, have not, never) , which help ensure the meaning of sentence is consistent with the label. 

\textbf{Overlapping Rate:} We calculate the overlapping rate between claim and evidence using Jaccard for unordered text overlapping rate and LCS index (Longest Common Sub-sequence) for ordered text overlapping. The results illustrated in Table \ref{tab:LLM-Generated Dataset Linguistics features Statistics Claim-Evidence} show that the models tend to have a high rewriting rate for the \textit{Supported} label, evidenced by the fact that the Jaccard rate and LCS index for the \textit{Supported} label are generally higher than those for the other two labels in the LLM-generated datasets. Additionally, it can be seen that Gemini, in the separate stages, exhibits high creativity, as indicated by lower Jaccard rates and LCS indices compared to the other models.

\section{MANUAL EVALUATION}
\label{subsec:MANUAL EVALUATION}
\subsection{Human Evaluation on Criteria}
\label{subsec:Human Evaluation On Criteria}

\begin{figure}[htb]
    \centering
    \includegraphics[width=1\textwidth]{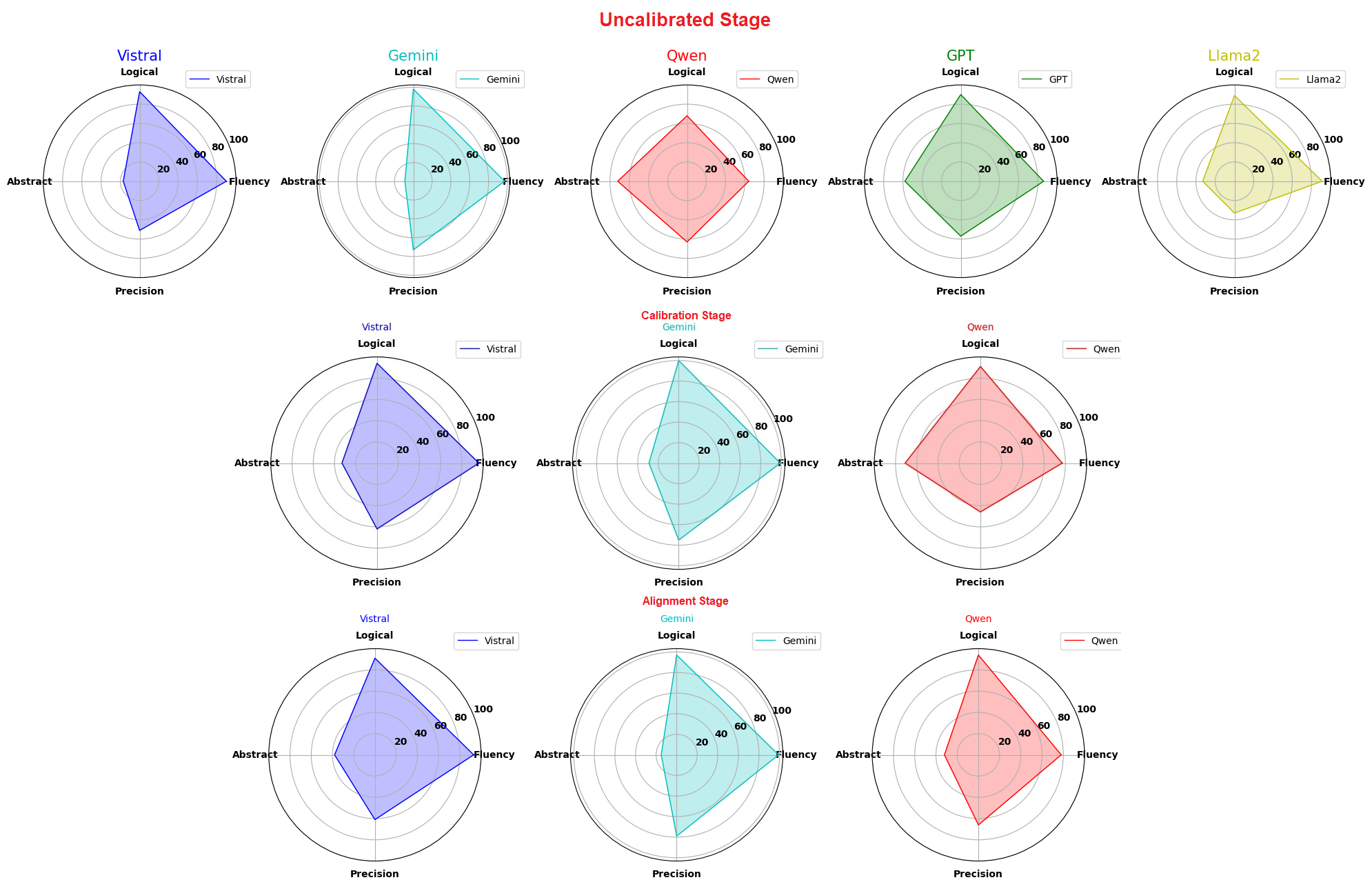} 
    \caption{Human evaluation in three stages of data generation (\%).} 
    \label{fig:human evaluation stage 1}
\end{figure}
To evaluate the quality of LLM-generated data, we conduct a manual assessment with 4 criteria. We take inspiration from \citeauthor{le2024lampat}, \citeyear{le2024lampat} for the two criteria, "fluency" and "logical", as they are suitable for automatic data generation. Additionally, we include two more criteria, "abstract" and "precision", to better fit the fact-checking task. We expected the claim to be general from evidence, which requires model inference, moreover, in this task, we generate the labeled data by generating a sentence not labeling classification so that it obligates precision in the generated sentence.

\begin{itemize}
    \item \textbf{Fluency:} Claim sentence is fluent and coherent.
    \item \textbf{Logical:} The information in the claim sentences is all logical.
    \item \textbf{Abstract:} The claim sentence synthesizes or combines pieces of information from multiple evidence sentences.
    \item \textbf{Precision:} The information in the claim is consistent with the information in the evidence and complies with the input label.
\end{itemize}


In addition, to ensure the reliability of the manual evaluation, we train and evaluate the agreement among five annotators who are native Vietnamese speakers. We measure the inter-annotator agreement using Fleiss $\kappa$ score, and the achieved result is 83.23\%. Then, we randomly select 100 samples from each LLM-generated dataset across three stages. The results of the manual evaluation are summarized in Figure \ref{fig:human evaluation stage 1}.

Overall, the generated claim sentences exhibit high fluency, with methodological improvements leading to more logical and coherent claims. However, the models face challenges in synthesizing information from multiple evidence sentences, as reflected in the "abstract" criterion. Additionally, the models show inconsistency in generating semantically accurate sentences for each label case, which is evident in the "precision" criterion, where most rates fall below 70\%. Qwen is uniform in all indicators whereas Gemini and Vistral tend to choose only one evidence sentence to create a claim. Llama 2 does very well on the Fluency and Logical criteria, but it does poorly on the Precision criteria.

\subsection{Examples of Assessment Criteria}
\label{subsec:Examples Of Assessment Criteria}
LLMs demonstrate impressive fluency and logical coherence in sentence production, marking significant progress in natural language processing. However, they sometimes struggle with typical sentence construction issues, such as lexical arrangement, resulting in sentences that lack smoothness and cause discomfort to the reader.
\begin{table}[htb]
\scriptsize
\centering
\begin{tabular}{p{0.2cm}|p{5.8cm}|p{4cm}}
\hline
\multicolumn{1}{p{1cm}|}{\textbf{Category}} & \multicolumn{1}{c|}{\textbf{Evidence}} & \multicolumn{1}{c}{\textbf{Claim}} \\ \hline
\multicolumn{1}{p{0.5cm}|}{\textbf{Correct Cases}} & Khí hậu Hà Nội mang đặc điểm của khí hậu nhiệt đới gió mùa, được nêu trên trang web chính thức của Hà Nội. Tuy nhiên, do chịu sự tác động mạnh mẽ của gió mùa nên thời gian bắt đầu và kết thúc của mỗi mùa thường không đồng đều nhau giữa các năm, nên sự phân chia các tháng chỉ mang tính tương đối.

(\textit{Hanoi's climate exhibits characteristics of a tropical monsoon climate, as stated on Hanoi's official website. However, due to the strong impact of the monsoons, the start and end times of each season often vary from year to year, making the division of months only relative.})
& Khí hậu Hà Nội được phân loại là khí hậu nhiệt đới gió mùa, nhưng do tác động của gió mùa nên thời gian bắt đầu và kết thúc của các mùa không đồng đều giữa các năm.

(\textit{Hanoi's climate is classified as a tropical monsoon climate, but due to the impact of monsoons, the start and end times of the seasons vary unevenly from year to year.}) \\ \hline
\multicolumn{1}{p{0.5cm}|}{\textbf{Wrong Cases}} & \textcolor{red}{Tất cả các ngôi nhà} hai bên đường khu phố cổ đều theo kiểu nhà ống, mang nét đặc trưng: bề ngang hẹp, chiều dài sâu, đôi khi thông sang phố khác. Trong khu 36 phố phường thuộc dự án bảo tồn, hiện chỉ còn một vài nhà cổ có giá trị, còn lại hầu hết đã được xây mới hoặc cải tạo tùy tiện.

(\textit{\textcolor{red}{All the houses} on both sides of the streets in the Old Quarter are tube houses, characterized by their narrow width and deep length, sometimes extending through to another street. In the 36 streets conservation project, only a few valuable old houses remain, while most have been newly built or renovated haphazardly.})
& \textcolor{red}{Một số các nhà} trong khu vực 36 phố phường dự án bảo tồn là nhà cổ có giá trị. 

(Temporary translation: \textit{\textcolor{red}{Some the houses} in the area of conservation project 36 streets are valuable ancient houses.})\\ \hline
\end{tabular}
\caption{Examples of correct and incorrect cases in sentence generation according to the fluency criterion (These claims are both \textit{\textit{Supported}} labels and generated by Gemini model).}
\label{tab: Fluency}
\end{table}

Table \ref{tab: Fluency} illustrates an example where an LLM generated a sentence that, despite being logically correct, is awkward for readers. The sentence contains two determiners, "Một số" (some) and "các" (the), before a noun, and lacks a preposition between two objects, "khu vực 36 phố phường" (the area of 36 streets) and "dự án bảo tồn" (conservation project). These issues make the sentence unnatural and difficult to read.\\



\begin{table}[htb]  
\scriptsize
\captionsetup{width=\linewidth, justification=centering}
\begin{tabular}{p{1cm}|p{5.8cm}|p{4cm}}
\hline
\multicolumn{1}{p{1cm}|}{\textbf{Category}} & \multicolumn{1}{c|}{\textbf{Evidence}} & \multicolumn{1}{c}{\textbf{Claim}} \\ \hline
\multicolumn{1}{p{0.5cm}|}{\textbf{Correct Cases}} & Trong IEEE 802.3, một datagram được gọi là một packet (gói) hoặc một frame. Phần giữa frame gồm có dữ liệu payload gồm bất cứ header nào dành cho các protocol khác (ví dụ, Internet Protocol) được mang trong frame.

(\textit{In IEEE 802.3, a datagram is referred to as a packet or a frame. The middle part of the frame consists of the data payload, including any headers for other protocols (e.g., Internet Protocol) that are carried within the frame.})
& Trong IEEE 802.3, phần giữa frame chứa dữ liệu payload và các header bổ sung từ các giao thức khác như Internet Protocol không được mang trong frame. 

(\textit{In IEEE 802.3, the middle part of the frame contains the data payload, and additional headers from other protocols, such as Internet Protocol, are not included within the frame.}) \\ \hline
\multicolumn{1}{p{0.5cm}|}{\textbf{Wrong Cases}} & Số lượng khách du lịch đến Việt Nam tăng nhanh nhất trong vòng \textcolor{red}{10 năm từ 2000–2010}. Năm 2013, có gần 7,6 triệu lượt khách quốc tế đến Việt Nam và năm 2017, có hơn 10 triệu lượt khách quốc tế đến Việt Nam, các thị trường lớn nhất là Trung Quốc, Hàn Quốc, Nhật Bản, Hoa Kỳ và Đài Loan.

(\textit{The number of tourists visiting Vietnam increased the fastest during the \textcolor{red}{10 years from 2000 to 2010}. In 2013, nearly 7.6 million international tourists visited Vietnam, and in 2017, over 10 million international tourists visited the country. The largest markets were China, South Korea, Japan, the United States, and Taiwan.})
& Số lượng khách du lịch đến Việt Nam tăng nhanh nhất trong vòng \textcolor{red}{10 năm từ 2000–2010 đến 2010–2020}.

(\textit{The number of tourists visiting Vietnam increased the fastest during the \textcolor{red}{10 years from 2000–2010 to 2010–2020}.})
\\ \hline
\end{tabular}
\caption{Examples of correct and incorrect cases in sentence generation according to the logical criterion \\(These claims are both \textit{Refuted} labels and generated by Vistral model).}
\label{tab: Logical}
\end{table}

\begin{table}[H]
\scriptsize  
\captionsetup{width=\linewidth, justification=centering}
\begin{tabular}{p{0.5cm}|p{5.8cm}|p{4cm}}
\hline
\multicolumn{1}{c|}{\textbf{Category}} & \multicolumn{1}{c|}{\textbf{Evidence}} & \multicolumn{1}{c}{\textbf{Claim}} \\ \hline
\multicolumn{1}{p{0.8cm}|}{\textbf{Correct Cases}} & Suốt nhiều thế kỷ ở Napoli có lệnh cấm tuyệt đối xây dựng các công trình kiến trúc mới bên ngoài các bức tường thành cho đến khi \textcolor{red}{phó vương Toledo đến thành phố vào thế kỷ 16} – đánh dấu bước ngoặt thay đổi lớn cho bộ mặt của Napoli, tiến hành \textcolor{red}{tái cấu trúc và mở rộng gấp đôi diện tích bề mặt đô thị} và kết nối cả 3 tòa lâu đài lại với nhau. \textcolor{red}{Điều này dẫn đến sau này, một số phần của các bức tường thành đã bị phá bỏ để giảm tắc nghẽn đô thị.}

(\textit{For centuries, there was an absolute ban on constructing new buildings outside the city walls of Naples until \textcolor{red}{Viceroy Toledo arrived in the 16th century} - his arrival marked a significant turning point for the city's appearance, \textcolor{red}{he undertook a restructuring and doubled the urban area}, connecting all three castles together. \textcolor{red}{This led to later on, some sections of the city walls being demolished to reduce urban congestion.}})
& Việc tái cấu trúc và mở rộng Napoli do phó vương Toledo thực hiện vào thế kỷ 16 đã dẫn đến việc phá hủy một số phần của các bức tường thành để giảm bớt tắc nghẽn đô thị. 

(\textit{The restructuring and expansion of Naples carried out by Viceroy Toledo in the 16th century led to the demolition of some sections of the city walls to alleviate urban congestion.}) \\ \hline
\multicolumn{1}{p{0.8cm}|}{\textbf{Wrong Cases}} & Vì những bất đồng về bản quyền kéo dài, The Beatles là một trong số những nghệ sĩ tên tuổi cuối cùng ký kết phân phối nhạc trực tuyến. \textcolor{red}{Năm 2010, toàn bộ 13 album phòng thu chính thức của The Beatles, cùng các album tuyển tập Past Masters, Album đỏ và Album xanh cuối cùng cũng được xuất hiện trên hệ thống phân phối của iTunes.}

(\textit{Due to prolonged copyright disputes, The Beatles were among the last major artists to sign up for online music distribution. \textcolor{red}{In 2010, all 13 of The Beatles' official studio albums, along with the Past Masters, Red Album, and Blue Album compilations, were finally made available on the iTunes distribution system.}})
& Album phòng thu chính thức của The Beatles đã được xuất hiện trên hệ thống phân phối của iTunes.

(\textit{The Beatles' official studio albums were made available on the iTunes distribution system.})
\\ \hline
\end{tabular}
\caption{Examples of correct and incorrect cases in sentence generation according to the abstract criterion \\(These claims are both \textit{\textit{Supported}} labels and generated by Gemini model).}
\label{tab: Examples of correct and incorrect cases in abstract}
\end{table}

Table \ref{tab: Logical} illustrates a logical error in LLM-generated claims. For instance, the model inaccurately compresses the time periods 2000-2010 and 2010-2020 into a total of 10 years, which is evidently incorrect. Such errors are particularly prevalent in \textit{Refuted} claims, where the model is required to deviate from the provided evidence. When tasked with generating information that contradicts or alters the original evidence, the model often struggles to maintain logical consistency. This example underscores the need for further refinement in prompt engineering and model training to better handle the complexities of generating logically sound refutations.

Most LLMs have the general tendency to choose only one piece of evidence to write a claim (Figure \ref{fig:human evaluation stage 1}). This is completely opposite when we want LLMs to be able to create a claim that contains information from multiple pieces of evidence (Table \ref{tab: Examples of correct and incorrect cases in abstract}).

LLMs also have cases whereas claims include information for which LLMs are trained, outside the scope of evidence, especially in evidence that does not explicitly mention subjects or locations. This information is not completely accurate and may be misleading. Table \ref{tab: Hallucinate} presents correct and incorrect cases using their own training data. The first sentence references Vietnam, but based solely on the evidence provided, the origin of these endangered rhinos remains unclear. The LLMs incorrectly identify Indonesia as the location in the claim sentence and even provide a reason for the rhinos' decline. In the second incorrect case, the evidence mentions an anonymous character, but the LLM generates that is "Zhang Yimou", despite the entire paragraph from which the evidence is extracted discussing "Jin Yong". 

\begin{table}[htb]
\scriptsize
\centering
\begin{tabular}{p{0.5cm}|p{5.8cm}|p{4cm}}
\hline
\multicolumn{1}{c|}{\textbf{Category}} & \multicolumn{1}{c|}{\textbf{Evidence}} & \multicolumn{1}{c}{\textbf{Claim}} \\ \hline
\multicolumn{1}{p{0.5cm}|}{\textbf{Wrong Cases}} & Dù được con người quan tâm bảo tồn nhưng diện tích cư trú của tê giác vẫn bị xâm lấn 15\% vào năm 1990 và con số tê giác còn sống đã giảm xuống chỉ còn 5-8 con vào năm 1999. Sau đó cơ quan bảo tồn ước tính chỉ còn 3-8 con tê giác, và có thể không có con đực nào sống sót.

(\textit{Despite conservation efforts, the habitat of rhinoceroses continued to be encroached upon, with a 15\% reduction in their living area by 1990. The number of surviving rhinoceroses dwindled to just 5-8 individuals by 1999. Subsequently, conservation agencies estimated that only 3-8 rhinoceroses remained, with the possibility of no surviving males.})
& Số lượng tê giác còn sống ở \textcolor{red}{Indonesia} đã giảm mạnh trong những năm qua, do diện tích cư trú bị thu hẹp và \textcolor{red}{nạn săn trộm}.

(\textit{The number of surviving rhinoceroses in \textcolor{red}{Indonesia} has drastically declined in recent years due to habitat loss and \textcolor{red}{poaching}.})
 \\ \hline
\multicolumn{1}{p{0.5cm}|}{\textbf{Wrong Cases}} & Năm 1946 từ biệt gia đình, ông về Hàng Châu làm phóng viên cho tờ Đông Nam nhật báo theo lời giới thiệu của Trần Hướng Bình, người ngày xưa đã tìm đến trường ông. Ông còn viết phê bình điện ảnh.

(\textit{In 1946, he bid farewell to his family and returned to Hangzhou to work as a reporter for the Dong Nam newspaper, following an introduction by Tran Huong Binh, someone who had once sought him out. He also wrote film reviews.})
& Năm 1946, \textcolor{red}{Trương Nghệ Mưu} trở thành phóng viên cho tờ Đông Nam nhật báo ở Hàng Châu, đồng thời viết phê bình điện ảnh.

(\textit{In 1946, \textcolor{red}{Zhang Yimou} became a reporter for the Dong Nam newspaper in Hangzhou, while also writing film reviews.})
 \\ \hline
\multicolumn{1}{p{0.5cm}|}{\textbf{Correct Cases}} & Vào năm 340, Alexandros lúc này mới 16 tuổi và đã hoàn thành xong quá trình học tập với Aristoteles, đã được vua cha triệu về Pella để hướng dẫn cách thức trị quốc. Các thành bang ngoại trừ Sparta đều chào đón quân đội Macedonia.

(\textit{In 340 BC, at the age of 16, Alexandros had just completed his studies with Aristotle and was summoned back to Pella by his father, the king, to learn about governing the kingdom. The city-states, excluding Sparta, welcomed the Macedonian army.})
& Vào năm 340 TCN, Alexandros đã trở thành \textcolor{red}{nhiếp chính} cho vua cha Philippos II, và các thành bang ngoại trừ Sparta đều chấp nhận quân đội Macedonia.

(\textit{In 340 BC, Alexandros became the \textcolor{red}{regent} for his father, King Philip II, and the city-states, excluding Sparta, all accepted the Macedonian army.})
 \\ \hline
\end{tabular}
\caption{An example of hallucination (The claims is all in \textit{\textit{Supported}} label and generated by model Vistral).}
\label{tab: Hallucinate}
\end{table}

However, the additional information provided by LLMs is not always inaccurate. In the third sentence, although the evidence does not mention that Alexandros would be a regent, the claim includes this information, which aligns with historical facts—Alexandros was indeed chosen as a regent when his father attacked Byzantion. Nevertheless, such hallucinations completely violate our instructions when creating \textit{\textit{Supported}} labels. These instances are more appropriate for generating a \textit{NEI} label, as they require models to combine pieces of information from the evidence with their own knowledge to create a claim sentence entirely irrelevant to the evidence. Despite adjustments to hyperparameters, hallucinations seem to occur randomly, presenting a persistent challenge when working with LLMs.

\section{AUTOMATIC EVALUATION}
\label{sec:AUTOMATIC EVALUATION}
\subsection{Baseline Models}
\label{subsec:Baseline Model}
For the task of verdict prediction, we select transformer-based language models for their robust performance across various languages and sentence structures. XLM-R \cite{xlmr} by Meta AI\footnote{https://ai.meta.com/} excels in cross-lingual understanding, is pre-trained on a diverse set of languages. mBERT of Google \cite{bert} captures cross-lingual relationships effectively, thanks to its extensive multilingual training data. mDeBERTa by Microsoft \cite{he2020deberta,he2022debertav3} improves upon BERT \cite{bert} by enhancing long-range dependencies and nuanced language understanding. PhoBERT \cite{nguyen-tuan-nguyen-2020-phobert}, from VinAI Research\footnote{https://www.vinai.io/}, is optimized for Vietnamese, leveraging extensive training on Vietnamese texts to excel in Vietnamese language processing.

\subsection{Prompting Experimental Results}
\label{subsec:Prompting Experimental Result}

\newcommand{\applyGradient}[1]{
    \ifdim #1 pt > 75 pt \cellcolor{gray!#1}#1
    \else\ifdim #1 pt > 50 pt \cellcolor{lightgray!#1}#1
    \else\cellcolor{white!#1}#1
    \fi\fi
}

To investigate the effectiveness of different large language models in generating data for Vietnamese fact-checking, we conducted experiments to evaluate model performance across distinct types of test sets. These test sets include one composed of data automatically generated by a large language model (LLM) and another curated by human experts, allowing us to examine how well models trained on LLM-generated data generalize to more complex, human-authored claims. The following results provide insights into the models' strengths and limitations, highlighting the challenges LLMs encounter when transitioning from synthetic to human-authored data.

\begin{table*}[htb]
\centering
\resizebox{\textwidth}{!}{%
\begin{tabular}{c||c|c|cc:cc:cc|cc:cc}
\cline{1-13}
\multicolumn{1}{c||}{\multirow{3}{*}{\textbf{Stage}}} & \multirow{3}{*}{\textbf{Model}} & \multirow{3}{*}{\textbf{Test Set}} & \multicolumn{6}{c|}{\textbf{Syllable based}} & \multicolumn{4}{c}{\textbf{Word based}} \\ \cline{4-13}
\multicolumn{1}{c||}{} &  &  & \multicolumn{2}{c:}{\textbf{XLM-R$_{Large}$}} & \multicolumn{2}{c:}{\textbf{mBERT}} & \multicolumn{2}{c|}{\textbf{mDeBERTa}} & \multicolumn{2}{c:}{\textbf{PhoBERT$_{Large}$}} & \multicolumn{2}{c}{\textbf{PhoBERT$_{V2}$}} \\ \cdashline{4-13}
\multicolumn{1}{c||}{} &  &  & F1 & Acc & F1 & Acc & F1 & Acc & F1 & Acc & F1 & Acc \\ \cline{1-13}
\multirow{10}{*}{\textbf{\begin{tabular}{c}
   Uncalibrated \\
   stage
\end{tabular}}} & \multirow{2}{*}{Vistral} & LLM & \applyGradient{76.36} & \applyGradient{76.17} & \applyGradient{72.84} & \applyGradient{72.72} & \applyGradient{76.57} & \applyGradient{76.42} & \applyGradient{74.33} & \applyGradient{74.26} & \applyGradient{74.95} & \applyGradient{74.75} \\ 
 &  & Human & \applyGradient{40.67} & \applyGradient{44.49} & \applyGradient{36.71} & \applyGradient{40.69} & \applyGradient{40.61} & \applyGradient{45.05} & \applyGradient{37.66} & \applyGradient{41.51} & \applyGradient{39.54} & \applyGradient{43.06} \\ \cdashline{2-13}
 & \multirow{2}{*}{GPT 3.5} & LLM & \applyGradient{75.47} & \applyGradient{75.35} & \applyGradient{72.51} & \applyGradient{72.33} & \applyGradient{73.08} & \applyGradient{73.01} & \applyGradient{74.04} & \applyGradient{73.93} & \applyGradient{75.22} & \applyGradient{75.10} \\
 &  & Human & \applyGradient{35.09} & \applyGradient{39.32} & \applyGradient{35.14} & \applyGradient{38.64} & \applyGradient{40.12} & \applyGradient{42.26} & \applyGradient{33.40} & \applyGradient{39.27} & \applyGradient{37.46} & \applyGradient{40.51} \\ \cdashline{2-13}
 & \multirow{2}{*}{Qwen} & LLM & \applyGradient{94.58} & \applyGradient{94.53} & \applyGradient{93.70} & \applyGradient{93.74} & \applyGradient{94.78} & \applyGradient{94.80} & \applyGradient{94.07} & \applyGradient{94.07} & \applyGradient{94.07} & \applyGradient{94.06} \\
 &  & Human & \applyGradient{39.85} & \applyGradient{43.80} & \applyGradient{37.09} & \applyGradient{39.82} & \applyGradient{40.02} & \applyGradient{43.25} & \applyGradient{34.44} & \applyGradient{38.58} & \applyGradient{33.78} & \applyGradient{39.95} \\ \cdashline{2-13}
 & \multirow{2}{*}{Llama 2} & LLM & \applyGradient{56.59} & \applyGradient{56.89} & \applyGradient{55.59} & \applyGradient{55.58} & \applyGradient{56.37} & \applyGradient{57.69} & \applyGradient{54.62} & \applyGradient{55.05} & \applyGradient{53.12} & \applyGradient{54.53} \\
 &  & Human & \applyGradient{33.99} & \applyGradient{39.63} & \applyGradient{33.41} & \applyGradient{37.83} & \applyGradient{32.11} & \applyGradient{39.39} & \applyGradient{35.83} & \applyGradient{38.27} & \applyGradient{37.35} & \applyGradient{38.51} \\ \cdashline{2-13}
 & \multirow{2}{*}{Gemini} & LLM & \applyGradient{69.36} & \applyGradient{69.53} & \applyGradient{63.79} & \applyGradient{64.04} & \applyGradient{66.81} & \applyGradient{66.79} & \applyGradient{66.47} & \applyGradient{66.78} & \applyGradient{65.19} & \applyGradient{65.93} \\
 &  & Human & \applyGradient{45.28} & \applyGradient{47.02} & \applyGradient{44.25} & \applyGradient{45.73} & \applyGradient{45.61} & \applyGradient{47.23} & \applyGradient{46.57} & \applyGradient{47.61} & \applyGradient{47.07} & \applyGradient{49.65} \\ \hline 
\multirow{6}{*}{\textbf{\begin{tabular}{c}
   Calibration \\
   stage
\end{tabular}}} & \multirow{2}{*}{Vistral} & LLM & \applyGradient{71.18} & \applyGradient{71.15} & \applyGradient{68.94} & \applyGradient{68.57} & \applyGradient{72.01} & \applyGradient{71.97} & \applyGradient{69.26} & \applyGradient{69.21} & \applyGradient{68.84} & \applyGradient{69.22} \\
 &  & Human & \applyGradient{46.23} & \applyGradient{46.61} & \applyGradient{41.83} & \applyGradient{41.87} & \applyGradient{47.68} & \applyGradient{48.60} & \applyGradient{44.18} & \applyGradient{44.18} & \applyGradient{47.59} & \applyGradient{49.04} \\ \cdashline{2-13}
 & \multirow{2}{*}{Gemini} & LLM & \applyGradient{72.92} & \applyGradient{73.05} & \applyGradient{67.82} & \applyGradient{67.95} & \applyGradient{73.29} & \applyGradient{73.61} & \applyGradient{70.16} & \applyGradient{70.36} & \applyGradient{70.45} & \applyGradient{70.73} \\
 &  & Human & \applyGradient{44.79} & \applyGradient{47.85} & \applyGradient{38.86} & \applyGradient{43.75} & \applyGradient{45.94} & \applyGradient{48.97} & \applyGradient{41.12} & \applyGradient{45.36} & \applyGradient{42.14} & \applyGradient{45.55} \\ \cdashline{2-13}
 & \multirow{2}{*}{Qwen} & LLM & \applyGradient{91.67} & \applyGradient{91.65} & \applyGradient{91.43} & \applyGradient{91.44} & \applyGradient{91.28} & \applyGradient{91.30} & \applyGradient{91.96} & \applyGradient{91.95} & \applyGradient{92.19} & \applyGradient{92.16} \\
 &  & Human & \applyGradient{26.07} & \applyGradient{37.21} & \applyGradient{22.42} & \applyGradient{35.28} & \applyGradient{27.81} & \applyGradient{37.46} & \applyGradient{20.88} & \applyGradient{34.85} & \applyGradient{24.68} & \applyGradient{35.72} \\ \hline
\end{tabular}%
}
\caption{The experimental results of the uncalibrated and calibration stages across all three labels.}
\label{tab:stage 1 2 experiment result}
\end{table*}

Table \ref{tab:stage 1 2 experiment result} illustrates the performance of various language models (LMs) on the data generation for fact-checking task, evaluated across two distinct test sets: one derived from datasets generated by an LLM and another curated by humans. Models consistently exhibit superior performance on the LLM-generated test sets compared to the human-curated ones. This disparity suggests that LMs, when trained on data created by other large language models, face challenges in evaluating claims from human-generated text, likely due to the greater diversity and complexity inherent in human language. Notably, the highest accuracy on the human-curated test set is a modest 49.65\%, achieved with data generated by Gemini and evaluated by PhoBERT.

Datasets created by LLMs such as Qwen, Vistral, or GPT-3.5 display good to very good performance on their respective test sets, with accuracies often exceeding 70\%. In particular, the Qwen dataset achieves an impressive accuracy of over 94\% on its test set, indicating that certain characteristics of LLM-generated datasets facilitate high performance. However, the performances on human test set underscore that LLM-generated datasets fail to capture the full complexity and variability of human language. Interestingly, the Gemini dataset, which only attains moderate accuracy on its own test set, achieves the highest accuracy on the human test set. This indicates that Gemini-generated data potentially embodies features of human language more closely than other LLMs.

In the calibration stage, we select high-performing LLMs for data generation: Gemini for API, Qwen for multilingual, and Vistral for Vietnamese monolingual models. Adjusting the prompt reduces performance on the machine-generated test set, especially for Vistral, but improves accuracy on the human-created test set, suggesting better generalizability to human content.

Performance varies by model as we can see in Table \ref{tab:stage 1 2 experiment result}, the dataset generated by Qwen shows poorer results on the human test set compared to the uncalibrated stage, with F1-score and accuracy dropping by over 10\%, indicating issues with label consistency. The dataset of Gemini improves slightly in syllable-based models but declines significantly in word-based models, highlighting different reactions to the same dataset. Vistral maintains high performance across evaluations, suggesting its datasets align well with evaluation models.

While LLMs are useful for generating extensive datasets, models trained mostly on LLM-generated data may struggle to generalize to the varied and noisier human-generated content. This underscores the need for incorporating diverse, human-generated data in training and evaluation to enhance model robustness and generalizability.

\section{SUPERVISED-FINETUNING LLMS}
\label{sec:Supervised-finetuning LLM}
\subsection{Why Do We Need To Fine-Tune LLMs in the Generated Task?}
\label{subsubsec:Why do we need to fine-tune LLMs in the generated task?}
The discrepancy between accuracy and F1-score on the human test set suggests the model struggles to differentiate between labels. We hypothesize that the \textit{NEI} label is particularly challenging. This label requires introducing or altering information to make it unrelated to the evidence, tasks that are complex and error-prone. Additionally, generating such content may conflict with LLMs' design to prioritize accuracy and truthfulness. Consequently, creating data for the \textit{NEI} label can lead to inconsistencies, highlighting the need for refined data generation and evaluation strategies to improve model reliability.

Table \ref{tab:2 label experiment} evaluates language models with just two labels (\textit{Supported} and \textit{Refuted}). The improved performance in this binary classification, compared to the three-label setup which includes \textit{NEI} labels, highlights the complexity of the \textit{NEI} label. The models show enhanced accuracy and F1-scores, with improvements up to 30\%, indicating clearer differentiation between \textit{Supported} and \textit{Refuted} claims. 
\begin{table}[htb]
\centering
\resizebox{\textwidth}{!}{%
\begin{tabular}{c||c|c|cccccc|cccc}
\cline{1-13}
\multicolumn{1}{c||}{\multirow{3}{*}{\textbf{Stage}}} & \multirow{3}{*}{\textbf{Model}} & \multicolumn{1}{c|}{} & \multicolumn{6}{c|}{\textbf{Syllable - Based}} & \multicolumn{4}{c}{\textbf{Word - Based}} \\ \cline{4-13}
\multicolumn{1}{c||}{} & \multicolumn{1}{c|}{} & \multicolumn{1}{c|}{} & \multicolumn{2}{c}{\textbf{XLM-R$_{Large}$}} & \multicolumn{2}{c}{\textbf{mBERT}} & \multicolumn{2}{c|}{\textbf{mDeBERTa}} & \multicolumn{2}{c}{\textbf{PhoBERT$_{Large}$}} & \multicolumn{2}{c}{\textbf{PhoBERT$_{V2}$}} \\ \cline{4-13}
\multicolumn{1}{c||}{} & & \multicolumn{1}{c|}{\multirow{-3}{*}{\textbf{Test Set}}} & F1 & Acc & F1 & Acc & F1 & \multicolumn{1}{c|}{Acc} & F1 & Acc & F1 & Acc \\ \cline{1-13}
\multicolumn{1}{c||}{} &  & LLM & \applyGradient{80.71} & \applyGradient{80.84} & \applyGradient{77.27} & \applyGradient{77.75} & \applyGradient{80.28} & \applyGradient{80.29} & \applyGradient{81.05} & \applyGradient{81.38} & \applyGradient{79.18} & \applyGradient{79.29} \\
\multicolumn{1}{c||}{} & \multirow{-2}{*}{Vistral} & Human & \applyGradient{67.77} & \applyGradient{68.37} & \applyGradient{62.78} & \applyGradient{62.80} & \applyGradient{63.93} & \applyGradient{65.58} & \applyGradient{62.85} & \applyGradient{62.98} & \applyGradient{58.14} & \applyGradient{60.20} \\ \cdashline{2-13}
\multicolumn{1}{c||}{} &  & LLM & \applyGradient{81.23} & \applyGradient{81.25} & \applyGradient{80.55} & \applyGradient{80.59} & \applyGradient{81.26} & \applyGradient{81.34} & \applyGradient{82.47} & \applyGradient{82.48} & \applyGradient{82.65} & \applyGradient{82.76} \\
\multicolumn{1}{c||}{} & \multirow{-2}{*}{GPT 3.5} & Human & \applyGradient{51.10} & \applyGradient{53.06} & \applyGradient{50.53} & \applyGradient{53.06} & \applyGradient{50.87} & \applyGradient{52.41} & \applyGradient{52.10} & \applyGradient{51.30} & \applyGradient{51.54} & \applyGradient{53.97} \\ \cdashline{2-13}
\multicolumn{1}{c||}{} &  & LLM & \applyGradient{96.77} & \applyGradient{96.77} & \applyGradient{95.15} & \applyGradient{95.15} & \applyGradient{96.16} & \applyGradient{96.16} & \applyGradient{95.45} & \applyGradient{95.45} & \applyGradient{96.67} & \applyGradient{96.67} \\
\multicolumn{1}{c||}{} & \multirow{-2}{*}{Qwen} & Human & \applyGradient{59.38} & \applyGradient{59.74} & \applyGradient{56.33} & \applyGradient{56.40} & \applyGradient{61.22} & \applyGradient{59.16} & \applyGradient{57.05} & \applyGradient{57.38} & \applyGradient{56.68} & \applyGradient{56.58} \\ \cdashline{2-13}
\multicolumn{1}{c||}{} &  & LLM & \applyGradient{63.22} & \applyGradient{63.32} & \applyGradient{63.62} & \applyGradient{63.68} & \applyGradient{66.39} & \applyGradient{66.43} & \applyGradient{65.72} & \applyGradient{66.61} & \applyGradient{68.39} & \applyGradient{68.51} \\
\multicolumn{1}{c||}{} & \multirow{-2}{*}{Llama 2} & Human & \applyGradient{49.60} & \applyGradient{54.73} & \applyGradient{49.23} & \applyGradient{52.13} & \applyGradient{39.58} & \applyGradient{51.11} & \applyGradient{42.48} & \applyGradient{49.44} & \applyGradient{40.05} & \applyGradient{49.81} \\ \cdashline{2-13}
\multicolumn{1}{c||}{} &  & LLM & \applyGradient{84.02} & \applyGradient{84.00} & \applyGradient{77.92} & \applyGradient{77.98} & \applyGradient{83.49} & \applyGradient{83.49} & \applyGradient{81.43} & \applyGradient{81.48} & \applyGradient{79.36} & \applyGradient{79.37} \\
\multicolumn{1}{c||}{\multirow{-10}{*}{\textbf{\begin{tabular}[c]{@{}c@{}}Uncalibrated stage \\ (2 labels)\end{tabular}}}} & \multirow{-2}{*}{Gemini} & Human & \applyGradient{71.07} & \applyGradient{71.24} & \applyGradient{60.38} & \applyGradient{61.50} & \applyGradient{73.50} & \applyGradient{73.65} & \applyGradient{67.06} & \applyGradient{67.07} & \applyGradient{64.79} & \applyGradient{65.02} \\ \cline{1-13}
\multicolumn{1}{c||}{} &  & LLM & \applyGradient{84.20} & \applyGradient{84.21} & \applyGradient{78.11} & \applyGradient{78.11} & \applyGradient{84.39} & \applyGradient{84.40} & \applyGradient{82.27} & \applyGradient{82.27} & \applyGradient{84.12} & \applyGradient{84.12} \\
\multicolumn{1}{c||}{} & \multirow{-2}{*}{Vistral} & Human & \applyGradient{71.51} & \applyGradient{71.52} & \applyGradient{63.48} & \applyGradient{63.50} & \applyGradient{66.59} & \applyGradient{66.98} & \applyGradient{65.33} & \applyGradient{65.68} & \applyGradient{71.22} & \applyGradient{71.24} \\ \cdashline{2-13}
\multicolumn{1}{c||}{} &  & LLM & \applyGradient{92.38} & \applyGradient{92.40} & \applyGradient{85.07} & \applyGradient{85.24} & \applyGradient{90.17} & \applyGradient{90.20} & \applyGradient{85.99} & \applyGradient{86.17} & \applyGradient{87.35} & \applyGradient{87.36} \\
\multicolumn{1}{c||}{} & \multirow{-2}{*}{Gemini} & Human & \applyGradient{75.01} & \applyGradient{75.04} & \applyGradient{65.02} & \applyGradient{65.40} & \applyGradient{75.32} & \applyGradient{75.32} & \applyGradient{69.52} & \applyGradient{70.04} & \applyGradient{68.48} & \applyGradient{68.73} \\ \cdashline{2-13}
\multicolumn{1}{c||}{} &  & LLM & \applyGradient{93.35} & \applyGradient{93.35} & \applyGradient{92.71} & \applyGradient{92.71} & \applyGradient{93.35} & \applyGradient{93.35} & \applyGradient{89.55} & \applyGradient{89.55} & \applyGradient{91.45} & \applyGradient{91.45} \\
\multicolumn{1}{c||}{\multirow{-6}{*}{\textbf{\begin{tabular}[c]{@{}c@{}}Calibration stage\\  (2 labels)\end{tabular}}}} & \multirow{-2}{*}{Qwen} & Human & \applyGradient{46.31} & \applyGradient{54.27} & \applyGradient{40.48} & \applyGradient{52.04} & \applyGradient{46.31} & \applyGradient{54.27} & \applyGradient{42.97} & \applyGradient{52.60} & \applyGradient{42.16} & \applyGradient{52.88} \\ \cline{1-13}
\end{tabular}
}
\caption{The experimental results of the uncalibrated stage and calibration stage in \textit{\textit{Supported}} and \textit{Refuted} labels.}
\label{tab:2 label experiment}
\end{table}
In the uncalibrated stage, datasets from GPT 3.5 and Llama 2 exhibit moderate performance on the human test set with two labels. The highest F1 scores achieved are 52.10\% by PhoBERT\(_{Large}\) for GPT 3.5 and 49.60\% by XLM-R\(_{Large}\) for Llama 2. In contrast, datasets from Gemini and Vistral yield significantly better performances, with mDeBERTa achieving an F1 score of 73.65\% and XLM-R\(_{Large}\) scoring 67.77\%.

In the calibration stage, both Gemini and Vistral show their improved performances, with F1 scores increasing significantly. Gemini achieves an impressive F1 score of 75.32\% using mDeBERTa, and Vistral reaches 71.51\% with XLM-R\(_{Large}\), highlighting their enhanced ability to handle the human test set with two labels. In contrast, Qwen experiences a substantial drop in F1 score, plummeting from 61.22\% to 46.31\%, indicating significant challenges in maintaining accuracy under the same conditions. The results of the 2-label human data set trained on data from LLMs models such as Vistral and Gemini are 15 - 30\% higher than the 3-label data set. We can therefore conclude the difficulty in creating \textit{NEI} labels of LLMs.

Removing the ambiguous \textit{NEI} label reduces confusion and boosts performance. Therefore, we choose to use Vietnamese fact-checking datasets to fine-tune our LLM models, aiming to enhance the \textit{NEI} label that suits our wishes as well as improve the performance of the other two labels.

\subsection{Results of Fine-Tuning Stage}
\label{subsubsec:Results Of Fine-Tuning Stage}
In the Supervised-Finetuning stage, we employ Vistral and Qwen using the SFT (Supervised Fine-Tuning) and LoRA (Low-Rank Adaptation) techniques \cite{lora}. For Gemini, we leverage Vertex AI\footnote{https://console.cloud.google.com/vertex-ai} of Google Cloud to fine-tune it with version 1.0-pro-002. The primary objective in this section is to enhance the performances of models on the \textit{NEI} label. Due to the limited availability of fact-checking datasets in Vietnamese, we utilize a single dataset for this task — ViWikiFC, comprising 20,916 claim sentences. We implement three distinct methods for this purpose:
\begin{itemize}
    \item \textbf{Synthetic Training}: this method involves training the model on a dataset containing all three labels (\textit{Supported}, \textit{Refuted}, \textit{NEI}). Trained model is then used to generate additional synthetic data.
    \item \textbf{Specific Training}: in this approach, we isolate the \textit{NEI} labels from the ViWikiFC dataset and fine-tune the models specifically on this subset. Subsequently, we generate claims labeled as \textit{NEI} and combine them with the two-label data (\textit{\textit{Supported}} and \textit{Refuted}) has the higher results between 2 stages.
    \item \textbf{Semi-Synthetic Training}: is a hybrid approach that combines data from the two-label dataset having the higher results between 2 stages with the \textit{NEI} data from the Synthetic Training method.
\end{itemize}


\begin{table*}[htb]
\centering
\resizebox{\textwidth}{!}{%
\begin{tabular}{c||c|c|cc:cc:cc|cc:cc}
\hline
\multirow{3}{*}{\textbf{Model}} & \multirow{3}{*}{\textbf{Method}} & \multirow{3}{*}{\textbf{Test Set}} & \multicolumn{6}{c}{\textbf{Syllable - Based}} & \multicolumn{4}{c}{\textbf{Word - Based}} \\ \cline{4-13}
 &  &  & \multicolumn{2}{c:}{\textbf{XLM-R$_{Large}$}} & \multicolumn{2}{c:}{\textbf{mBERT}} & \multicolumn{2}{c|}{\textbf{mDeBERTa}} & \multicolumn{2}{c:}{\textbf{PhoBERT$_{Large}$}} & \multicolumn{2}{c}{\textbf{PhoBERT$_{V2}$}} \\ \cdashline{4-13}
 &  &  & F1 & Acc & F1 & Acc & F1 & Acc & F1 & Acc & F1 & Acc \\ \hline
\multirow{6}{*}{Vistral} & \multirow{2}{*}{Synthetic training} & LLM & \applyGradient{69.28} & \applyGradient{69.51} & \applyGradient{64.90} & \applyGradient{65.30} & \applyGradient{69.13} & \applyGradient{69.39} & \applyGradient{65.58} & \applyGradient{65.79} & \applyGradient{66.68} & \applyGradient{66.46} \\ 
 &  & Human & \applyGradient{49.39} & \applyGradient{50.96} & \applyGradient{47.47} & \applyGradient{48.03} & \applyGradient{53.12} & \applyGradient{53.82} & \applyGradient{51.01} & \applyGradient{51.52} & \applyGradient{50.77} & \applyGradient{51.46} \\ \cdashline{2-13}
 & \multirow{2}{*}{\begin{tabular}[c]{@{}c@{}}Semi synthetic \\  training\end{tabular}} & LLM & \applyGradient{85.82} & \applyGradient{85.67} & \applyGradient{80.22} & \applyGradient{80.52} & \applyGradient{86.53} & \applyGradient{86.62} & \applyGradient{86.15} & \applyGradient{86.30} & \applyGradient{85.78} & \applyGradient{85.75} \\
 &  & Human & \applyGradient{47.41} & \applyGradient{48.10} & \applyGradient{43.80} & \applyGradient{45.73} & \applyGradient{48.58} & \applyGradient{48.66} & \applyGradient{46.77} & \applyGradient{47.17} & \applyGradient{49.15} & \applyGradient{49.47} \\ \cdashline{2-13}
 & \multirow{2}{*}{\begin{tabular}[c]{@{}c@{}} Specific training\end{tabular}} & LLM & \applyGradient{85.97} & \applyGradient{86.06} & \applyGradient{82.34} & \applyGradient{82.28} & \applyGradient{84.41} & \applyGradient{84.44} & \applyGradient{81.72} & \applyGradient{81.76} & \applyGradient{83.97} & \applyGradient{83.88} \\
 &  & Human & \applyGradient{51.89} & \applyGradient{51.77} & \applyGradient{45.64} & \applyGradient{47.98} & \applyGradient{52.07} & \applyGradient{52.08} & \applyGradient{48.84} & \applyGradient{48.85} & \applyGradient{48.26} & \applyGradient{49.16} \\ \hline
\multirow{6}{*}{Gemini} & \multirow{2}{*}{Synthetic training} & LLM & \applyGradient{82.39} & \applyGradient{82.26} & \applyGradient{77.63} & \applyGradient{77.50} & \applyGradient{82.06} & \applyGradient{82.01} & \applyGradient{80.22} & \applyGradient{80.18} & \applyGradient{81.14} & \applyGradient{81.04} \\
 &  & Human & \applyGradient{55.87} & \applyGradient{55.32} & \applyGradient{38.39} & \applyGradient{44.31} & \applyGradient{51.29} & \applyGradient{53.39} & \applyGradient{38.03} & \applyGradient{43.92} & \applyGradient{43.74} & \applyGradient{47.67} \\ \cdashline{2-13}
 & \multirow{2}{*}{\begin{tabular}[c]{@{}c@{}}Semi synthetic \\  training\end{tabular}} & LLM & \applyGradient{84.25} & \applyGradient{84.30} & \applyGradient{77.54} & \applyGradient{77.46} & \applyGradient{83.19} & \applyGradient{83.32} & \applyGradient{83.16} & \applyGradient{83.20} & \applyGradient{83.26} & \applyGradient{83.26} \\
 &  & Human & \applyGradient{43.38} & \applyGradient{47.60} & \applyGradient{36.43} & \applyGradient{42.63} & \applyGradient{45.65} & \applyGradient{48.60} & \applyGradient{40.67} & \applyGradient{45.86} & \applyGradient{41.35} & \applyGradient{46.11} \\ \cdashline{2-13}
 & \multirow{2}{*}{\begin{tabular}[c]{@{}c@{}} Specific training\end{tabular}} & LLM & \applyGradient{88.74} & \applyGradient{88.70} & \applyGradient{82.34} & \applyGradient{82.28} & \applyGradient{87.97} & \applyGradient{87.97} & \applyGradient{87.21} & \applyGradient{87.04} & \applyGradient{87.12} & \applyGradient{87.05} \\
 &  & Human & \applyGradient{50.48} & \applyGradient{52.02} & \applyGradient{45.64} & \applyGradient{47.98} & \applyGradient{50.89} & \applyGradient{52.27} & \applyGradient{54.01} & \applyGradient{55.19} & \applyGradient{51.82} & \applyGradient{52.33} \\ \cdashline{2-13}
\hline
\multirow{6}{*}{Qwen} & \multirow{2}{*}{Synthetic training} & LLM & \applyGradient{75.03} & \applyGradient{75.37} & \applyGradient{72.34} & \applyGradient{72.42} & \applyGradient{74.93} & \applyGradient{75.19} & \applyGradient{75.48} & \applyGradient{75.55} & \applyGradient{75.16} & \applyGradient{75.33} \\
 &  & Human & \applyGradient{49.67} & \applyGradient{51.27} & \applyGradient{44.16} & \applyGradient{46.23} & \applyGradient{49.95} & \applyGradient{51.34} & \applyGradient{51.32} & \applyGradient{52.17} & \applyGradient{50.96} & \applyGradient{52.21} \\ \cdashline{2-13}
 & \multirow{2}{*}{\begin{tabular}[c]{@{}c@{}}Semi synthetic \\  training\end{tabular}} & LLM & \applyGradient{96.83} & \applyGradient{96.62} & \applyGradient{96.12} & \applyGradient{96.18} & \applyGradient{97.11} & \applyGradient{97.07} & \applyGradient{95.49} & \applyGradient{95.44} & \applyGradient{96.62} & \applyGradient{96.60} \\
 &  & Human & \applyGradient{39.04} & \applyGradient{41.93} & \applyGradient{36.16} & \applyGradient{38.83} & \applyGradient{38.93} & \applyGradient{42.13} & \applyGradient{42.20} & \applyGradient{43.93} & \applyGradient{34.28} & \applyGradient{39.70} \\ \cdashline{2-13}
 & \multirow{2}{*}{\begin{tabular}[c]{@{}c@{}} Specific training\end{tabular}} & LLM & \applyGradient{97.28} & \applyGradient{97.31} & \applyGradient{94.55} & \applyGradient{94.47} & \applyGradient{96.14} & \applyGradient{96.20} & \applyGradient{94.21} & \applyGradient{94.15} & \applyGradient{96.02} & \applyGradient{95.94} \\
 &  & Human & \applyGradient{36.17} & \applyGradient{39.02} & \applyGradient{36.53} & \applyGradient{39.01} & \applyGradient{36.29} & \applyGradient{39.21} & \applyGradient{36.01} & \applyGradient{40.82} & \applyGradient{36.76} & \applyGradient{39.95} \\ \cdashline{2-13}
\hline
\end{tabular}%
}
\caption{The experimental results of the alignment stage fine-tuning method}
\label{tab:stage 3 experiment result}
\end{table*}

Table \ref{tab:stage 3 experiment result} shows the results of three methods. The results are generally improved compared to the uncalibrated stage and the calibration stage but not completely superior (approximately 6\% higher for PhoBERT$_{V2}$ in Specific training of Gemini compared to mDeBERTa of Gemini data in the calibration stage). Overall, generating data using the Synthetic Training method generally yields better performance compared to previous stages, although it still produces relatively low results on the human dataset. Notably, in several cases, the Specific Training method achieves results nearly equivalent to Synthetic Training (Gemini). The Semi-Synthetic method performed the worst, with no results of models exceeding 50\% on the human dataset.

\subsection{Performances on ViNLI Test Set}
\label{subsec:Performances On ViNLI Test Set}
We see that most LLMs tend to use only one sentence from the evidence to create a claim, so we decide to test their performances on ViNLI \cite{huynh-etal-2022-vinli}, a dataset also use one sentence to generate another to see if they obtain higher results. The results in Table \ref{tab:vinli experiment} show the performance of various LLMs on the ViNLI test set. In the uncalibrated stage, Llama2 and GPT-3.5 demonstrate moderate performance across all models, with Llama2 achieving F1 scores ranging from 35.36\% (mBERT) to 38.17\% (XLM-R\(_{Large}\)) and GPT-3.5 slightly outperforming it with scores between 36.79\% (mBERT) and 39.88\% (mDeBERTa). However, Gemini and Vistral surpass both Llama2 and GPT-3.5, with Vistral showing the highest performance in most metrics, notably achieving an F1 score of 42.88\% (PhoBERT\(_{Large}\)). In the calibration stage, a general decline in performance is observed for Qwen compared to the uncalibrated stage. Still, for Gemini and Vistral, the results are significantly enhanced, achieving an F1 score of 56.32\% (mDeBERTa - Vistral).

\newcommand{\applyGradientvnli}[1]{
    \ifdim #1 pt > 60 pt \cellcolor{darkgray!#1}#1
    \else\ifdim #1 pt > 50 pt \cellcolor{gray!#1}#1
    \else\cellcolor{white!#1}#1
    \fi\fi
}
\begin{table*}[htb]
\centering
\resizebox{\textwidth}{!}{%
\begin{tabular}{c||l|cccccc|cccc}
\hline
\multirow{3}{*}{\textbf{Stage}} & \multicolumn{1}{c|}{\multirow{3}{*}{\textbf{LLM}}} & \multicolumn{6}{c|}{\textbf{Syllable - Based}} & \multicolumn{4}{c}{\textbf{Word - Based}} \\ 
\cline{3-12}
 & \multicolumn{1}{c|}{} & \multicolumn{2}{c}{\textbf{mBERT}} & \multicolumn{2}{c}{\textbf{mDeBERTa}} & \multicolumn{2}{c|}{\textbf{XLM-R$_{Large}$}} & \multicolumn{2}{c}{\textbf{PhoBERT$_{Base}$}} & \multicolumn{2}{c}{\textbf{PhoBERT$_{Large}$}} \\ 
\cline{3-12}
 & \multicolumn{1}{c|}{} & F1 & Acc & F1 & Acc & F1 & Acc & F1 & Acc & F1 & Acc \\ 
\hline
\multirow{5}{*}{\begin{tabular}{c}
   Uncalibrated stage
\end{tabular}} & Llama2 & \applyGradientvnli{35.36} & \applyGradientvnli{40.06} & \applyGradientvnli{36.37} & \applyGradientvnli{44.26} & \applyGradientvnli{38.17} & \applyGradientvnli{43.92} & \applyGradientvnli{35.87} & \applyGradientvnli{43.73} & \applyGradientvnli{34.12} & \applyGradientvnli{42.31} \\ 
\cdashline{2-12}
 & Qwen & \applyGradientvnli{31.34} & \applyGradientvnli{37.46} & \applyGradientvnli{46.56} & \applyGradientvnli{48.81} & \applyGradientvnli{45.61} & \applyGradientvnli{49.23} & \applyGradientvnli{39.11} & \applyGradientvnli{43.24} & \applyGradientvnli{38.82} & \applyGradientvnli{44.57} \\ 
\cdashline{2-12}
 & GPT3.5 & \applyGradientvnli{36.79} & \applyGradientvnli{43.02} & \applyGradientvnli{39.88} & \applyGradientvnli{45.58} & \applyGradientvnli{39.32} & \applyGradientvnli{45.69} & \applyGradientvnli{38.64} & \applyGradientvnli{44.79} & \applyGradientvnli{39.46} & \applyGradientvnli{45.53} \\ 
\cdashline{2-12}
 & Gemini & \applyGradientvnli{40.23} & \applyGradientvnli{44.83} & \applyGradientvnli{43.32} & \applyGradientvnli{48.01} & \applyGradientvnli{42.98} & \applyGradientvnli{47.79} & \applyGradientvnli{44.21} & \applyGradientvnli{48.01} & \applyGradientvnli{46.39} & \applyGradientvnli{49.16} \\ 
\cdashline{2-12}
 & Vistral & \applyGradientvnli{39.46} & \applyGradientvnli{40.42} & \applyGradientvnli{41.82} & \applyGradientvnli{48.05} & \applyGradientvnli{41.93} & \applyGradientvnli{47.49} & \applyGradientvnli{41.73} & \applyGradientvnli{46.82} & \applyGradientvnli{42.88} & \applyGradientvnli{47.97} \\ 
\hline
\multirow{3}{*}{\begin{tabular}{c}
    Calibration stage
\end{tabular}} & Qwen & \applyGradientvnli{29.06} & \applyGradientvnli{38.21} & \applyGradientvnli{37.56} & \applyGradientvnli{44.04} & \applyGradientvnli{38.52} & \applyGradientvnli{44.91} & \applyGradientvnli{29.48} & \applyGradientvnli{39.40} & \applyGradientvnli{32.51} & \applyGradientvnli{40.28} \\ 
\cdashline{2-12}
 & Gemini & \applyGradientvnli{43.44} & \applyGradientvnli{46.47} & \applyGradientvnli{44.75} & \applyGradientvnli{50.49} & \applyGradientvnli{49.61} & \applyGradientvnli{53.01} & \applyGradientvnli{40.47} & \applyGradientvnli{46.55} & \applyGradientvnli{43.76} & \applyGradientvnli{48.19} \\ 
\cdashline{2-12}
 & Vistral & \applyGradientvnli{43.89} & \applyGradientvnli{44.35} & \applyGradientvnli{56.32} & \applyGradientvnli{56.47} & \applyGradientvnli{55.13} & \applyGradientvnli{55.42} & \applyGradientvnli{53.22} & \applyGradientvnli{52.96} & \applyGradientvnli{51.24} & \applyGradientvnli{51.50} \\ 
\hline
\multirow{3}{*}{\begin{tabular}[c]{@{}c@{}}Alignment stage\\  (Sythetic Training)\end{tabular}} & Qwen & \applyGradientvnli{50.6} & \applyGradientvnli{51.19} & \applyGradientvnli{59.39} & \applyGradientvnli{60.12} & \applyGradientvnli{59.02} & \applyGradientvnli{60.03} & \applyGradientvnli{55.46} & \applyGradientvnli{55.91} & \applyGradientvnli{55.07} & \applyGradientvnli{56.01} \\ 
\cdashline{2-12}
 & Gemini & \applyGradientvnli{42.35} & \applyGradientvnli{47.66} & \applyGradientvnli{54.09} & \applyGradientvnli{56.40} & \applyGradientvnli{54.52} & \applyGradientvnli{56.71} & \applyGradientvnli{46.35} & \applyGradientvnli{50.88} & \applyGradientvnli{45.04} & \applyGradientvnli{50.32} \\ 
\cdashline{2-12}
 & Vistral & \applyGradientvnli{48.87} & \applyGradientvnli{49.87} & \applyGradientvnli{59.15} & \applyGradientvnli{59.23} & \applyGradientvnli{58.92} & \applyGradientvnli{59.10} & \applyGradientvnli{53.58} & \applyGradientvnli{54.24} & \applyGradientvnli{55.8} & \applyGradientvnli{55.92} \\ 
\hline
\multirow{3}{*}{\begin{tabular}[c]{@{}c@{}}Alignment stage\\  (Semi Sythetic Training)\end{tabular}} & Qwen & \applyGradientvnli{34.1} & \applyGradientvnli{41.96} & \applyGradientvnli{40.02} & \applyGradientvnli{45.54} & \applyGradientvnli{41.23} & \applyGradientvnli{46.75} & \applyGradientvnli{35.27} & \applyGradientvnli{42.97} & \applyGradientvnli{43.61} & \applyGradientvnli{46.55} \\ 
\cdashline{2-12}
 & Gemini & \applyGradientvnli{36.64} & \applyGradientvnli{44.13} & \applyGradientvnli{44.13} & \applyGradientvnli{49.34} & \applyGradientvnli{43.55} & \applyGradientvnli{48.86} & \applyGradientvnli{41.93} & \applyGradientvnli{47.43} & \applyGradientvnli{50.09} & \applyGradientvnli{53.01} \\ 
\cdashline{2-12}
 & Vistral & \applyGradientvnli{50.20} & \applyGradientvnli{51.41} & \applyGradientvnli{60.02} & \applyGradientvnli{60.06} & \applyGradientvnli{58.93} & \applyGradientvnli{59.68} & \applyGradientvnli{52.73} & \applyGradientvnli{54.15} & \applyGradientvnli{54.19} & \applyGradientvnli{55.27} \\ 
\hline 
\multirow{3}{*}{\begin{tabular}[c]{@{}c@{}}Alignment stage\\  (Specific Training)\end{tabular}} & Qwen & \applyGradientvnli{37.66} & \applyGradientvnli{42.66} & \applyGradientvnli{33.32} & \applyGradientvnli{42.23} & \applyGradientvnli{36.24} & \applyGradientvnli{42.31} & \applyGradientvnli{35.27} & \applyGradientvnli{42.98} & \applyGradientvnli{36.92} & \applyGradientvnli{42.80} \\ 
\cdashline{2-12}
 & Gemini & \applyGradientvnli{38.61} & \applyGradientvnli{45.54} & \applyGradientvnli{50.38} & \applyGradientvnli{53.26} & \applyGradientvnli{51.22} & \applyGradientvnli{52.46} & \applyGradientvnli{46.44} & \applyGradientvnli{50.48} & \applyGradientvnli{47.23} & \applyGradientvnli{49.27} \\ 
\cdashline{2-12}
 & Vistral & \applyGradientvnli{45.38} & \applyGradientvnli{48.54} & \applyGradientvnli{54.59} & \applyGradientvnli{55.83} & \applyGradientvnli{53.87} & \applyGradientvnli{54.59} & \applyGradientvnli{49.58} & \applyGradientvnli{51.90} & \applyGradientvnli{52.73} & \applyGradientvnli{54.15} \\
\hline
\end{tabular}}
\caption{The performance on ViNLI test set when using data generated by LLM of all stages.}
\label{tab:vinli experiment}
\end{table*}

In the alignment stage, with synthetic training, Qwen shows a significant improvement, especially in mDeBERTa and XLM-R\(_{Large}\), achieving an impressive F1 score of 59.39\% (mDeBERTa). Vistral also demonstrates notable performance, with F1 scores reaching 59.15\% (mDeBERTa). The semi-synthetic training stage generally increases performance for all models, with Vistral achieving the highest F1 scores, up to 60.02\% (mDeBERTa). However, the performance of Qwen drops compared to its synthetic training stage, although it remains competitive. In the specific training stage, Qwen and Vistral show varied performance, with Vistral maintaining relatively higher scores, while Gemini demonstrates the lowest performance in this stage across most models. 

Overall, testing on the ViNLI dataset reveals that models tend to perform better when using synthetic data generation methods. But in general, even using a data set with only 1 evidence sentence and one claim sentence, the accuracy of language models when trained on the dataset created by LLMs still gives results far worse than when training on the dataset made by humans.

\subsection{Why Cannot Fine-Tuning LLMs Achieve High Performances on Human Datasets?}
According to our hypothesis, large language models still lag behind human capabilities. Fine-tuning LLMs may not significantly enhance their performance, as demonstrated in \cite{tian2023fine}. Instead, fine-tuning primarily aids in better alignment with specific datasets, without inherently improving the model intelligence.To enhance LLM capabilities effectively, alternative approaches are required. Retraining LLMs with one or three labels can improve alignment with datasets but does not inherently boost the intelligence of model. A more promising strategy involves semi-manually generating data, particularly for small datasets, and subsequently scaling up generation for larger datasets.

\citeauthor{zhang2024careful} argued that large language models are trained in a competitive manner, prioritizing the achievement of high performance across all tasks within the available datasets, rather than fostering the ability to generalize knowledge. The Mistral model (the backbone of the Vistral model) ranks second, very close to the top-performing model in this contender-style training. This may explain why there is not a significant breakthrough in performance between supervised tuning and the calibration stage.

\section{CONCLUSION AND FUTURE WORK}
\label{sec:CONCLUSION AND FUTURE WORK}
In conclusion, this paper evaluates the data generation capabilities of large language models (LLMs) in the Vietnamese fact-checking task by assessing the performance of language models alongside manual evaluations. We also analyze the datasets generated by LLMs through various linguistic features. The findings reveal a notable disparity between the capabilities of LLMs and those of humans, particularly when the complexity of LLM-generated data diverges significantly from that of manually constructed data. Errors in the automatic data generation process contribute to the low performance of language models on human test sets. Furthermore, fine-tuning methods do not lead to substantial improvements in LLMs' ability to replicate human behavior, as evidenced by only minor changes in language model performance. Although the data generation capabilities of LLMs still encounter many limitations, their potential for automatic data generation remains generally promising. Nonetheless, enhanced methods are necessary to stabilize the data generation capabilities of LLMs and improve their quality.

Based on the findings and conclusions of this study, we gain several directions for future work that could enhance the data generation capabilities and performance of large language models (LLMs) in Vietnamese fact-checking tasks. First, we can integrate multi-dimensional evaluation criteria that account for the unique linguistic features of Vietnamese could significantly improve the quality and relevance of automatically generated data. By incorporating metrics for sentence structure, vocabulary, and semantics, LLMs can reduce the occurrence of unintended errors during data generation. Additionally, the option of applying Reinforcement Learning with Human Feedback (RLHF) also offers a promising method to align LLM outputs with human standards, helping models better understand accuracy requirements and verification standards. Finally, a deeper analysis of specific error types, such as grammatical, semantic, and structural mistakes, could also yield insights into areas for improvement, enabling more targeted model refinements.
\\
\section*{LIMITATIONS}
\label{sec:LIMITATIONS}
Although datasets constructed by LLMs incur significantly lower costs compared to manually constructed datasets (approximately 20 times lower than the ViNLI dataset \cite{huynh-etal-2022-vinli} when using Vistral) and also require less construction time, the analyses above indicate that the quality of the data is far from human performance, especially the results in Section \ref{subsec:MANUAL EVALUATION} which show instability in tasks involving semantic constraints. Moreover, we only use simple methods such as prompting with instruction and supervised fine-tuning, so the application of more complex methods \cite{rafailov2024direct} in automatic data generation is left for the future.

We analyze LLM capability through five large language models: Qwen, Gemini, Llama2, GPT3.5, and Vistral. Since Vistral\footnote{https://huggingface.co/Viet-Mistral/Vistral-7B-Chat} is a large language model designed for Vietnamese with only one version, the 7B parameters chat model, it is convenient for analysis and experimentation on large language models. For LLMs like Qwen\footnote{https://huggingface.co/Qwen/Qwen1.5-7B-Chat} and Llama2\footnote{https://huggingface.co/meta-llama/Llama-2-7b-chat}, we chose the 7B parameters chat version; for GPT, we used GPT3.5 turbo\footnote{https://platform.openai.com/docs/models/gpt-3-5-turbo}; for Gemini, we used the Gemini 1.0 pro version\footnote{https://console.cloud.google.com/vertex-ai/publishers/google/model-garden/gemini-pro?pli=1}. Our research focuses on the most basic cases by utilizing the basic versions of LLMs. However, we cannot determine whether higher parameter versions of LLMs with more complex training techniques offer better performance. Additionally, using more advanced versions requires increased resources, making it a problem that requires careful consideration in research on automatic data generation through LLMs.

\section*{ETHICS STATEMENT}
In this study, we evaluate the performance of Large Language Models (LLMs) on the specific task of Vietnamese fact-checking by automatically constructing a dedicated dataset using open-source information from Wikipedia. Our research is guided by the principles of responsible AI usage, aiming to enhance fact-checking capabilities for low-resource languages. We declare no conflicts of interest and commit to complete transparency in our methods and findings. To support reproducibility and further research, all generated data will be made publicly accessible.

\section*{Acknowledgement}
This research was supported by The VNUHCM-University of Information Technology's Scientific Research Support Fund of the VNUHCM University of Information Technology.

\section*{Declaration of Interest Statement}
The authors declare that they have no conflict of interest.

\section*{CRediT authorship contribution statement}
 \textbf{Long Truong To:} Conceptualization; Formal analysis; Investigation; Methodology; Validation; Visualization; Writing - review \& editing. \textbf{Hung Tuan Le:} Conceptualization; Data curation; Formal analysis; Investigation; Validation; Visualization; Writing - original draft. \textbf{Dat Van-Thanh Nguyen:} Conceptualization; Data curation; Investigation; Methodology; Writing - original draft. \textbf{Manh Trong Nguyen:} Data curation; Formal analysis; Investigation; Validation. \textbf{Tri Thien Nguyen:} Data curation; Investigation; Methodology. \textbf{Tin Van Huynh:} Conceptualization; Formal analysis; Investigation; Methodology; Validation; Supervision; Writing - review \& editing. \textbf{Kiet Van Nguyen:} Conceptualization; Formal analysis; Investigation; Methodology; Validation; Supervision; Writing - review \& editing.

\section*{Data Available}
Data will be made available on request.

\bibliographystyle{elsarticle-harv} 

\bibliography{cite}

\pagebreak

\section*{Appendix}
\appendix

\section{Prompt Sample}
We have 3 versions of the prompt corresponding to 3 different stages in the process of building data with LLMs. Our prompt will basically have 4 parts for the first 2 data generation stages: defining the role of LLMs in the problem, defining the concepts of the elements in the problem - here defining only the label we want LLMs to give claim in stage 1 (Table \ref{tab:stage-1-prompt} and all 3 labels in stage 2 (Table \ref{tab:stage-2-prompt}), defining the work that LLMs need to do, and finally examples. For the fine-tuning stage, the prompt does not have examples (Table \ref{tab:stage-3-prompt}). All the prompts are originally in Vietnamese but are translated into English in the tables below.

\begin{table*}[t]
    \centering
    \footnotesize
    \resizebox{\linewidth}{!}{
    \begin{tabular}{p{\linewidth}}
    \hline
[ROLE] You are an expert in the field of natural language processing, especially in the task of fact-checking.
\\
Your task is to create a CLAIM sentence with the SUPPORTED label in Vietnamese based on the information in EVIDENCE in the problem of fact-checking with the following instructions:
\\
1. The SUPPORTED label is the label determined when the information in the CLAIM sentence is created correctly with the information in EVIDENCE.
\\
2. You can infer new information but still have to ensure that the new information is correct and limited to the scope of information in EVIDENCE.
\\
3. [IMPORTANT NOTE] Do not copy exactly 1 sentence from the Evidence sentences.
\\
4. [IMPORTANT NOTE] No matter how many sentences there are in Evidence, you must combine the information in all the Evidence sentences provided when creating the Claim.
\\
5. [IMPORTANT NOTE] Just give me the Claim and nothing else.

$-------------------------------$

Here are some examples:
\\

$[$EVIDENCE$]$: The filmmakers began location scouting around April 2016; principal photography began on July 19 of that year and lasted four months until November 22. Wright and Hoeks's cut scene was one of the first scenes shot on set.

[Claim]: The scene between Wright and Hoeks was filmed in 2016.

\\
$[$EVIDENCE$]$: Although Romance of the Three Kingdoms contains some fictional historical details, in general, Chinese official histories also acknowledge that the Shu Han court had many commendable characters: the Shu Han king Liu Bei was originally from a humble background, and as a child he had to weave straw sandals to make a living, so he understood the suffering of the people very well. He built his empire from nothing with the loyal help of his generals, and when he ascended the throne, he implemented a policy of tolerance towards the people. Therefore, folk tales about the Three Kingdoms period tend to praise Liu Bei and the Shu Han dynasty, hating his enemies is inevitable, and the tendency to "support Liu and oppose Cao Cao" has been the common thought of the majority of Chinese people since before Romance of the Three Kingdoms was published.

[Claim]: Romance of the Three Kingdoms tends to favor the Shu and create hostility toward the Cao Dynasty.

\\
$[$EVIDENCE$]$: Early influences on the band include Elvis Presley, Carl Perkins, Little Richard, and Chuck Berry. About Presley, Lennon said: "Nothing attracted me until I heard Elvis."

[Claim]: Lennon expressed his admiration for Elvis when he heard him sing.

\\
$[$EVIDENCE$]$: As a famous leader in Southeast Asia, according to Clark D. Neher, Ho Chi Minh combined Marxism-Leninism with Vietnamese nationalism. The main ideology in Ho Chi Minh's struggles was to combine national liberation revolution with proletarian revolution.

[Claim]: The main ideology in Ho Chi Minh was to combine national liberation revolution with proletarian revolution, which originated from the combination of Marxism-Leninism with Vietnamese nationalism.

\\
$[$EVIDENCE$]$: The terrain of Da Lat is divided into two distinct types: mountainous terrain and mountainous plain terrain. To the south, the mountainous terrain transitions to a lower terrain level, characterized by the Prenn Pass area with high mountain ranges interspersed with deep valleys.

[Claim]: The mountainous terrain in Da Lat gradually decreases towards the South, especially the Prenn Pass area.\\

$-------------------------------$

Here are the EVIDENCE and CLAIM SUPPORTED labels you need to rely on to create CLAIM: \\ 

$[$EVIDENCE$]$: \{EVIDENCE\} \\

$[$CLAIM$]$: \{CLAIM\} \\
\hline
    \end{tabular}
    }
    
    \caption{Stage 1 - Uncalibrated prompt for \textit{Supported} label}
    \label{tab:stage-1-prompt}
\end{table*}

\clearpage
\begin{table}[htbp]
\small
\centering
\begin{tabular}{p{11cm}}
\hline
$[$ROLE$]$ You are an expert in the field of natural language processing, especially in the task of fact-checking. The Fact-Checking problem includes 3 labels SUPPORTED, REFUTED, NOT ENOUGH INFORMATION defined as follows: 

SUPPORTED$:$ is the information in the CLAIM sentence is correct based on the information in the EVIDENCE sentences. \vspace{1mm}

REFUTED: is the information in the CLAIM sentence is incorrect based on the information in the EVIDENCE sentence. \vspace{1mm}\\
NOT ENOUGH INFORMATION: is the information in the CLAIM sentence cannot be determined to be correct or incorrect based on the information in the EVIDENCE sentences. \vspace{2mm}\\

Your task is to create a CLAIM sentence with the REFUTED label in Vietnamese based on a SAMPLE CLAIM sentence with the SUPPORTED label, and a group of EVIDENCE sentences in the Fact-Checking problem with the following instructions: \vspace{2mm}\\

1. The REFUTED label is the label determined when the information in the CLAIM sentence is created FALSE, or contrary to the information in EVIDENCE. \vspace{1mm}\\

2. You can infer new information, but you must ensure that the new information is FALSE or contradictory and is limited to the information in EVIDENCE. \vspace{1mm}\\

3. [IMPORTANT NOTE] Do not copy the exact same sentence from the EVIDENCE statements. \vspace{1mm}\\

4. [IMPORTANT NOTE] You must combine the information in all the EVIDENCE statements provided when creating the CLAIM. \vspace{1mm}\\

5. [IMPORTANT NOTE] Show me the CLAIM and provide nothing else.\vspace{2mm}\\

$-------------------------------$ \vspace{2mm}\\

Here are some examples: \vspace{2mm}\\

$[$EVIDENCE$]$: The filmmakers began location scouting around April 2016; principal photography on the film began on July 19 of that year and lasted four months until November 22. The scene where Wright and Hoeks interjected was one of the first scenes shot on set.\\
$[$CLAIM SUPPORTED$]$: The film took about 4 months and 4 days for principal photography, and the first scene to be filmed was Wright and Hoeks's splicing scene.\\
$[$CLAIM REFUTED$]$: Wright and Hoeks's splicing scene was the last scene to be filmed on set, and it took only 2 months for principal photography.\\ \\

$[$EVIDENCE$]$: Although Romance of the Three Kingdoms has some fictional historical details, in general, Chinese official histories also acknowledge that the Shu Han Dynasty had many commendable characters: the Shu Han Dynasty's king Liu Bei was originally from a humble background, and had to weave straw sandals to make a living as a child, so he understood the suffering of the people very well. He built his empire from nothing with the loyal support of his generals, and when he ascended the throne, he implemented a lenient policy towards the people. Therefore, folk tales about the Three Kingdoms period tend to praise Liu Bei and the Shu Han Dynasty, hating his enemies is inevitable, and the tendency to $"$support Liu and oppose Cao$"$ has been the common thought of the majority of Chinese people since before the work Romance of the Three Kingdoms was born. \\
$[$CLAIM SUPPORTED$]$: Romance of the Three Kingdoms is a work that contains real characters in Chinese history, and the representative is the Shu Han Dynasty king Liu Bei.\\
\hline
\end{tabular}
\end{table}
\clearpage 

\begin{table}[htbp]
\small
\centering
\begin{tabular}{p{11cm}}
\hline
$[$CLAIM REFUTED$]$: Romance of the Three Kingdoms is a work that does not contain any characters at all. \\ 

$[$EVIDENCE$]$: Early influences on the band include Elvis Presley, Carl Perkins, Little Richard, and Chuck Berry. About Presley, Lennon said: "Nothing attracted me until I heard Elvis.\\
$[$CLAIM SUPPORTED$]$: Carl Perkins, Little Richard and Chuck Berry were musical figures and early influences on the band.\\
$[$CLAIM REFUTED$]$: Only Chuck Berry was musically active and an early influence on the band.\\ \\

$[$EVIDENCE$]$: The United States is a multicultural country, home to many diverse groups of races, traditions, and values. Dynamism is a characteristic of Americans, they always have the need to act to achieve their goals.\\
$[$CLAIM SUPPORTED$]$: The United States, also known as America, is a country with diversity in many typical aspects such as racial diversity.\\
$[$CLAIM REFUTED$]$: The United States is a country with low cultural diversity when only one race lives.\\ \\

$[$EVIDENCE$]$: Is a famous leader In Southeast Asia, according to Clark D. Neher, Ho Chi Minh combined Marxism-Leninism with Vietnamese nationalism. The main ideology in Ho Chi Minh's struggles was to combine national liberation revolution with proletarian revolution.\\
$[$CLAIM SUPPORTED$]$: Ho Chi Minh, a famous leader whose ideology was the combination of national liberation revolution with proletarian revolution.\\
$[$CLAIM REFUTED$]$: Ho Chi Minh, a famous leader in Europe when he only used the Marxist-Leninist school of thought as the main ideology of his struggles.
\\

$-------------------------------$

Here are the EVIDENCE and CLAIM SUPPORTED labels you need to rely on to create CLAIM: \\ 

$[$EVIDENCE$]$: \{EVIDENCE\} \\

$[$CLAIM SUPPORTED$]$: \{CLAIM SUPPORTED\} \\

$[$CLAIM REFUTED$]$:
\\ 
\hline
\end{tabular}
\caption{Stage 2 - Calibration prompt for \textit{Refuted} label} \label{tab:stage-2-prompt}
\end{table}

\begin{table*}[t]
\footnotesize
\centering
\begin{tabular}{p{\linewidth}}
\hline
\textbf{system}: 
$[$ROLE$]$ You are an expert in the field of natural language processing, especially in the task of fact-checking. The Fact-Checking problem includes 3 labels SUPPORTED, REFUTED, NOT ENOUGH INFORMATION defined as follows: \vspace{1mm}\\
SUPPORTED$:$ is the information in the CLAIM sentence is correct based on the information in the EVIDENCE sentences. \vspace{1mm}\\
REFUTED: is the information in the CLAIM sentence is incorrect based on the information in the EVIDENCE sentence. \vspace{1mm}\\
NOT ENOUGH INFORMATION: is the information in the CLAIM sentence cannot be determined to be correct or incorrect based on the information in the EVIDENCE sentences. \vspace{1mm}\\

Your task is to create a CLAIM sentence with the REFUTED label in Vietnamese based on a group of EVIDENCE sentences in the Fact-Checking problem with the following instructions: \vspace{2mm}\\

1. The REFUTED label is the label determined when the information in the CLAIM sentence is created FALSE, or contrary to the information in EVIDENCE. \vspace{1mm}\\

2. You can infer new information, but you must ensure that the new information is FALSE or contradictory and is limited to the information in EVIDENCE. \vspace{1mm}\\

3. [IMPORTANT NOTE] Do not copy the exact same sentence from the EVIDENCE statements. \vspace{1mm}\\

4. [IMPORTANT NOTE] You must combine the information in all the EVIDENCE statements provided when creating the CLAIM. \vspace{1mm}\\

5. [IMPORTANT NOTE] Show me the CLAIM and provide nothing else.\vspace{2mm}\\
\\
\textbf{user}:
[EVIDENCE]: $\{$EVIDENCE$\}$
\\
\\
\textbf{assistant}:
[CLAIM]: 
\\
\hline
    \end{tabular}
    \caption{Stage 3 - Alignment prompt for \textit{Refuted} label}
    \label{tab:stage-3-prompt}
\end{table*}

\end{document}